\documentclass[10pt,journal,compsoc]{IEEEtran}
\ifCLASSOPTIONcompsoc
  \usepackage[nocompress]{cite}
\else
  \usepackage{cite}
\fi
\ifCLASSINFOpdf
  \usepackage[pdftex]{graphicx}
\else
\fi
\usepackage{amsmath}
\usepackage{dsfont}
\usepackage{algorithmic}
\usepackage[linesnumbered,ruled,vlined]{algorithm2e}
\usepackage{amsmath,bm}
\usepackage{wrapfig}

\SetKwInput{KwInput}{Input}

\usepackage{xcolor}
\newcommand{\jenny}[1]{{\textcolor{blue}{Jenny: #1}}}

\newcommand{\lau}[1]{\textcolor{magenta}{Laura: #1}}

\usepackage{hyperref}

\ifCLASSOPTIONcompsoc
  \usepackage[caption=false,font=footnotesize,labelfont=sf,textfont=sf]{subfig}
\else
  \usepackage[caption=false,font=footnotesize]{subfig}
\fi
\begin{document}
%
\title{The Group Loss++: A deeper look into group loss for deep metric learning}
%
%
%
%

\author{Ismail Elezi*,~\IEEEmembership{}
        Jenny Seidenschwarz*,~\IEEEmembership{}
        Laurin Wagner*,~\IEEEmembership{}
        Sebastiano Vascon,~\IEEEmembership{}
        Alessandro Torcinovich,~\IEEEmembership{}
        Marcello Pelillo,~\IEEEmembership{IEEE Fellow,}
        and Laura Leal-Taixé~\IEEEmembership{}
\IEEEcompsocitemizethanks{\IEEEcompsocthanksitem I.E, J.S, L.W and L.L.T are with Dynamic Vision and Learning Group at the Technical University of Munich, S.V, A.T and M.P are at Ca' Foscari University of Venice\protect\\
Corresponding authors: Ismail Elezi (\url{ismail.elezi@tum.de}), Laura Leal-Taixé (leal.taixe@tum.de)
\IEEEcompsocthanksitem * denotes equal contribution}
}
\IEEEtitleabstractindextext{%
\begin{abstract}
Deep metric learning has yielded impressive results in tasks such as clustering and image retrieval by leveraging neural networks to obtain highly discriminative feature embeddings, which can be used to group samples into different classes.
Much research has been devoted to the design of smart loss functions or data mining strategies for training such networks. 
Most methods consider only pairs or triplets of samples within a mini-batch to compute the loss function, which is commonly based on the distance between embeddings. 
We propose {\it Group Loss}, a loss function based on a differentiable label-propagation method that enforces embedding similarity across {\it all} samples of a group while promoting, at the same time, low-density regions amongst data points belonging to different groups. 
Guided by the smoothness assumption that ``similar objects should belong to the same group'', the proposed loss trains the neural network for a classification task, enforcing a consistent labelling amongst samples within a class. We design a set of inference strategies tailored towards our algorithm, named \textit{Group Loss++} that further improve the results of our model. We show state-of-the-art results on clustering and image retrieval on four retrieval datasets, and present competitive results on two person re-identification datasets, providing a unified framework for retrieval and re-identification. 

\end{abstract}

}

\maketitle

\IEEEdisplaynontitleabstractindextext

%
\IEEEpeerreviewmaketitle

\IEEEraisesectionheading{\section{Introduction}\label{sec:introduction}}
\IEEEPARstart{M}easuring object similarity is at the core of many important machine learning problems like clustering and object retrieval. 
For visual tasks, this means learning a distance function over images. With the rise of deep neural networks, the focus has rather shifted towards learning a feature embedding that is easily separable using a simple distance function, such as the Euclidean distance. 
In essence, objects of the same class (similar) should be close by in the learned manifold, while objects of a different class (dissimilar) should be far away.

Historically, the best performing approaches get deep feature embeddings from the so-called siamese networks \cite{bromley1994signature}, which are typically trained using the contrastive loss \cite{bromley1994signature} or the triplet loss \cite{DBLP:conf/nips/SchultzJ03,DBLP:journals/jmlr/WeinbergerS09}. 
A clear drawback of these losses is that they only consider pairs or triplets of data points, missing key information about the relationships between all members of the mini-batch. On a mini-batch of size $n$, despite that the number of pairwise relations between samples is $\mathcal{O}(n^2)$, contrastive loss uses only $\mathcal{O}(n/2)$ pairwise relations, while triplet loss uses $\mathcal{O}(2n/3)$ relations.
Additionally, these methods consider only the relations between objects of the same class (positives) and objects of other classes (negatives), without making any distinction that negatives belong to different classes.
This leads to not taking into consideration the global structure of the embedding space, and consequently results in lower clustering and retrieval performance. 
To compensate for that, researchers rely on other tricks to train neural networks for deep metric learning: intelligent sampling \cite{DBLP:conf/iccv/ManmathaWSK17}, multi-task learning \cite{DBLP:conf/cvpr/ZhangZLZ16} or hard-negative mining \cite{DBLP:conf/cvpr/SchroffKP15}. 
Recently, researchers have been increasingly working towards exploiting in a principled way the global structure of the embedding space \cite{DBLP:journals/corr/abs-1906-07589,DBLP:conf/cvpr/Cakir0XKS19,DBLP:conf/cvpr/0003CBS18,DBLP:conf/cvpr/WangHKHGR19}, typically by designing ranking loss functions instead of following the classic triplet formulations.

In a similar spirit, we propose {\it Group Loss}, a novel loss function for deep metric learning that considers the similarity between all samples in a mini-batch. To create the mini-batch, we sample from a fixed number of classes, with samples coming from a class forming a \textit{group}. Thus, each mini-batch consists of several randomly chosen groups, and each group has a fixed number of samples. An iterative, fully-differentiable label propagation algorithm is then used to build feature embeddings which are similar for samples belonging to the same group, and dissimilar otherwise. 

At the core of our method lies an iterative process called {replicator dynamics} \cite{weibull1997evolutionary,DBLP:journals/neco/ErdemP12}, that refines the local information, given by the softmax layer of a neural network, with the global information of the mini-batch given by the similarity between embeddings. 
The driving rationale is that the more similar two samples are, the more they affect each other in choosing their final label and tend to be grouped together in the same group, while dissimilar samples do not affect each other on their choices.

We then study the embedding space generated by networks trained with our Group Loss, resulting in a few observations that we exploit by introducing a set of inference strategies. We call this model, the \textit{Group Loss++} and show that reaches significantly better results than the {\it Group Loss}, making clustering and image retrieval easier. Finally, we show that our proposed model can be used to train networks in the field of person re-identification, thus providing a similarity learning unified framework that works both for retrieval and re-identification. 

\begin{figure*}[!t]
\centering
\includegraphics[width=\textwidth, trim={0 4.9cm 0.2cm 0},clip]{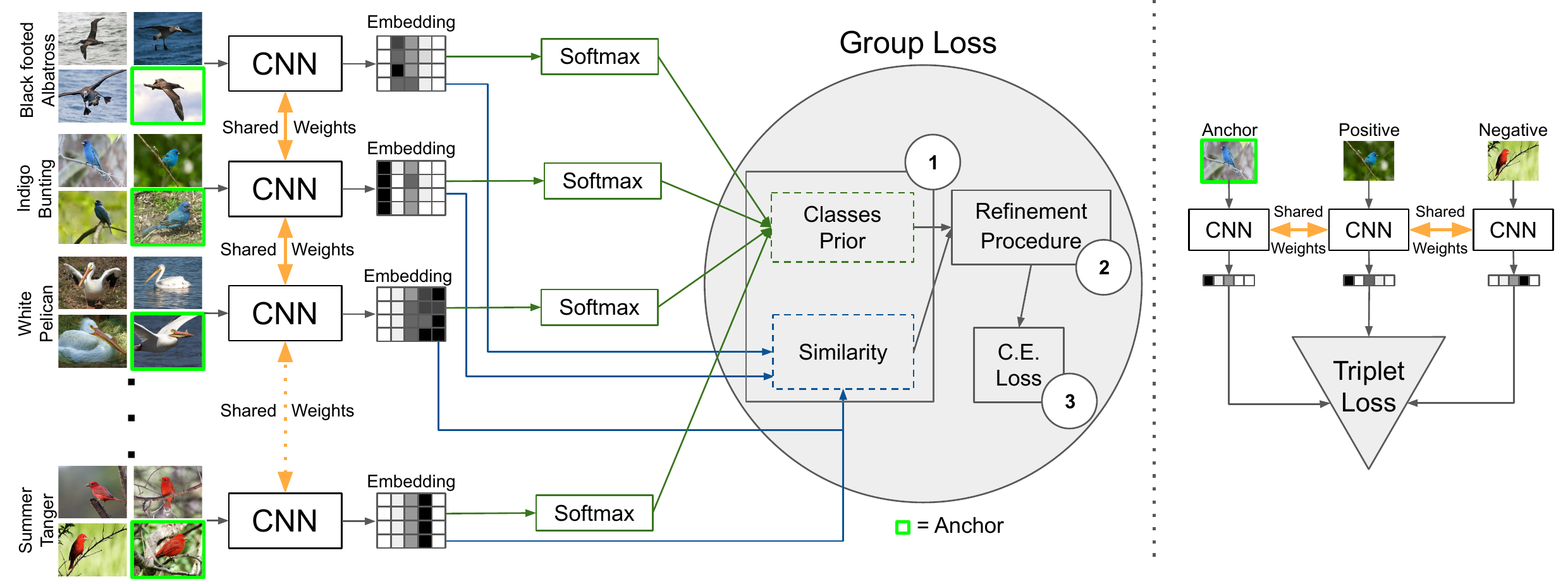}
\caption{A comparison between a neural model trained with the Group Loss (left) and the triplet loss (right). Given a mini-batch of images belonging to different classes, their embeddings are computed through a convolutional neural network. Such embeddings are then used to generate a similarity matrix that is fed to the Group Loss along with prior distributions of the images on the possible classes. The green contours around some mini-batch images refer to \emph{anchors}. It is worth noting that, differently from the triplet loss, the Group Loss considers multiple classes and the pairwise relations between all the samples. Numbers from \raisebox{.4pt}{\textcircled{\raisebox{-1pt} {1}}} to \raisebox{.4pt}{\textcircled{\raisebox{-1pt} {3}}} refer to the Group Loss steps, see Sec \ref{sec:gl_overview} for the details.}
\label{fig:pipeline}
\end{figure*}

Our \textbf{contribution} in this work is five-fold: 
\begin{itemize}
    \item We propose the \textit{Group Loss}, a novel loss function to train neural networks for deep metric embedding that takes into account the local information of the samples, as well as their similarity.
    \item We propose a differentiable label-propagation iterative model to embed the similarity computation within backpropagation, allowing end-to-end training with our new loss function.
    \item We introduce a set of inference strategies, resulting in \textit{Group Loss++} that improve the results of our model.
    \item We show state-of-the-art qualitative and quantitative results in four standard clustering and retrieval datasets. 
    \item We show competitive results on two person re-identification datasets, thus providing a unified framework for similarity learning. 
\end{itemize}
\section{Related Work}

\textbf{Classical metric learning losses.} The first attempt at using a neural network for feature embedding was done in the seminal work of Siamese Networks \cite{bromley1994signature}. A cost function called \textit{contrastive loss} was designed in such a way as to minimize the distance between pairs of images belonging to the same cluster and maximize the distance between pairs of images coming from different clusters. In \cite{DBLP:conf/cvpr/ChopraHL05}, researchers used the principle to successfully address the problem of face verification. 
Another line of research on convex approaches for metric learning led to the triplet loss \cite{DBLP:conf/nips/SchultzJ03,DBLP:journals/jmlr/WeinbergerS09}, which was later combined with the expressive power of neural networks \cite{DBLP:conf/cvpr/SchroffKP15}. The main difference from the original Siamese network is that the loss is computed using triplets (an anchor, a positive, and a negative data point). 
The loss is defined to make the distance between features of the anchor and the positive sample smaller than the distance between the anchor and the negative sample. The approach was so successful in the field of face recognition and clustering, that soon many works followed. The majority of works on the Siamese architecture consist of finding better cost functions, resulting in better performances on clustering and retrieval. In \cite{DBLP:conf/nips/Sohn16}, the authors generalized the concept of triplet by allowing a joint comparison among $N - 1$ negative examples instead of just one. 
\cite{DBLP:conf/cvpr/SongXJS16} designed an algorithm for taking advantage of the mini-batches during the training process by lifting the vector of pairwise distances within the batch to the matrix of pairwise distances, thus enabling the algorithm to learn feature embedding by optimizing a novel structured prediction objective on the lifted problem. The work was later extended in \cite{DBLP:conf/cvpr/SongJR017}, proposing a new metric learning scheme based on structured prediction that is designed to optimize a clustering quality metric, i.e., the normalized mutual information \cite{DBLP:journals/corr/abs-1110-2515}. Better results were achieved on \cite{DBLP:conf/iccv/WangZWLL17}, where the authors proposed a novel angular loss, which takes angle relationship into account. A very different problem formulation was given by \cite{DBLP:conf/icml/LawUZ17}, where the authors used a spectral clustering-inspired approach to achieve deep embedding. Similar to that is \cite{DBLP:conf/annpr/MeierEADS18} where the authors design a network to directly learn to cluster objects. A recent work presents several extensions of the triplet loss that reduce the bias in triplet selection by correcting the distribution shift on the selected triplets \cite{DBLP:conf/eccv/YuLGDT18}. Our work differs from all the mentioned works. While the goal is the same, finding good feature embeddings for every image, we do not use triplets but instead use all the relations in a minibatch, thus learning the global manifold of the dataset.


\textbf{Classification-based losses.}  Recently, the work of \cite{DBLP:journals/corr/abs-1811-12649} showed that a carefully tuned normalized softmax cross-entropy loss function combined with a balanced sampling strategy can achieve competitive results. A similar line of research is that of \cite{DBLP:conf/aaai/ZhengJSZWH19}, where the authors use a combination of normalized-scale layers and Gram-Schmidt optimization to achieve efficient usage of the softmax cross-entropy loss for metric learning. 
CenterLoss \cite{DBLP:conf/eccv/WenZL016} modifies the cross-entropy loss by simultaneously learning a center for deep
features of each class and penalizing the distances between the deep features and their corresponding class centers.
SoftTriple loss \cite{DBLP:journals/corr/abs-1909-05235} goes a step further by taking into consideration the similarity between classes. Furthermore, it uses multiple centers for class, allowing them to reach state-of-the-art results, at a cost of significantly increasing the number of parameters of the model. 
Our work significantly differs from the mentioned works. Instead of learning one or more centers, or considering the similarity between classes, we instead refine the features of each image based on the similarity with the other images in the dataset. By doing so, we achieve state-of-the-art results without increasing the number of parameters of the model.

\textbf{Proxy-losses.}
Related to SoftTriple loss \cite{DBLP:journals/corr/abs-1909-05235} are the proxy-losses.
The authors of \cite{DBLP:conf/iccv/Movshovitz-Attias17, DBLP:journals/corr/abs-2004-01113, DBLP:conf/cvpr/KimKCK20} proposed to optimize the triplet loss on a different space of triplets than the original samples, consisting of an anchor data point and similar and dissimilar learned proxy data points. These proxies approximate the original data points so that a triplet loss over the proxies is a tight upper bound of the original loss. The final formulation of the loss is shown to be similar to that of softmax cross-entropy loss, challenging the long-hold belief that classification losses are not suitable for the task of metric learning. Proxy-GML \cite{DBLP:journals/corr/abs-2010-13636} goes a step further by using reverse label propagation to adjust the neighbor relationships of the proxies based on their ground truth. While, like the mentioned works, our work modifies cross-entropy loss, we do not use any proxy representation. Furthermore, unlike Proxy-GML, at the heart of our work is a label propagation between the samples in the mini-batch, thus directly optimizing the embedding space.

\textbf{Person re-identification.} 
Similar to deep metric learning, Siamese Networks \cite{bromley1994signature} are historically utilized to generate feature embeddings since the the earliest deep learning-based works for person re-identification \cite{DBLP:conf/cvpr/LiZXW14, DBLP:conf/cvpr/AhmedJM15}.
Typically, works on person re-identification use versions of the triplet loss together with sampling strategies \cite{DBLP:conf/cvpr/ChenCZH17, DBLP:conf/eccv/GeHDS18, DBLP:journals/corr/abs-1711-08184, DBLP:journals/corr/HermansBL17, DBLP:conf/iccv/QuanDWZY19, DBLP:conf/cvpr/0004GLL019, DBLP:conf/iccv/DaiCGZT19}. 
Several works \cite{DBLP:conf/cvpr/ChengGZWZ16, DBLP:conf/eccv/SunZYTW18, DBLP:conf/eccv/VariorHW16, DBLP:conf/iccv/DaiCGZT19, DBLP:conf/cvpr/ZhangLZ019}, propose to subdivide the image of the whole person into several parts and process them
separately to enforce the architectures to learn local, more fine-grained features \cite{DBLP:conf/cvpr/ChengGZWZ16}. 
While many works rely on basic classification backbones such as ResNets \cite{7780459} or DenseNets \cite{DBLP:conf/cvpr/HuangLMW17}, several works introduce their own bakcbone specifically tailored to the task of person re-identification. For example, \cite{DBLP:conf/iccv/QuanDWZY19} proposed a neural architecture search
approach inspired by \cite{DBLP:conf/iclr/LiuSY19} that aims to find an optimal architecture by searching a
search space of operations to construct and concatenate neural cells in an automated way. 
Probably the closest to our work is that of \cite{DBLP:journals/corr/abs-1904-11397} where the authors target the task of person re-identification from a graph-theoretical point of view. Given
query and gallery images form an undirected graph and the images belonging to the same
identity should form a cluster. These clusters can be found by making use of constrained
dominant sets \cite{DBLP:conf/icpr/ZemeneAP16} in order to control the global shape of the energy landscape. 
The optimization of the clustering objective is done by a similar iterative procedure that uses replicator dynamics \cite{weibull1997evolutionary}. Unlike \cite{DBLP:journals/corr/abs-1904-11397}, we use the iterative procedure to spread the label preferences from images to their neighbors.


\section{Group Loss}

\subsection{Preliminaries}

Given a bunch of samples $I_1, I_2,...,I_i$, the task of metric learning is to learn a function that maps them to a compact representation (an embedding) $\phi(I_1), \phi(I_2),...,\phi(I_i)$  such that the representation of the samples coming from the same class should be close, while those coming from different classes should be further away. In inference, this function is used to find representations of samples that come from different classes, and then this representation is used to do retrieval (e.g. KNN) or clustering (e.g. K-Means \cite{MacQueen}). For the purpose of this work, we will consider that samples are images, although in principle, they can be any type of data.

The classical way of doing metric learning is to use \textbf{the contrastive} loss \cite{bromley1994signature} that minimizes the distance between pairs of images  belonging to the same class and maximizes the distance between pairs of images coming from different classes. Given two images $A$ and $B$, the loss is described by the following equation:

\begin{equation}\label{contrastive}
\begin{split}
    L(A, B) = y^* ||\phi(A) - \phi(B)||^2 + \\ (1 - y^*) max (0, m^2 - ||\phi(A) - \phi(B)||^2)
\end{split}
\end{equation}

where $y^*$ is an indicator variable that is $1$ if $A$ and $B$ belong to the same class and $0$ otherwise, while $m$ is a margin.

An extension of the contrastive loss is the \textbf{triplet loss} \cite{DBLP:conf/cvpr/SchroffKP15}, which takes as input an anchor $A$, a positive $P$, and a negative $N$, where the loss optimies the distance between $A$ and $P$ to be smaller (up to a margin $m$) than the distance between $A$ and $N$:

\begin{equation}\label{triplet}
\begin{split}
    L(A, P, N) = max(0, ||\phi(A) - \phi(P)||^2 - \\ ||\phi(A) - \phi(N)||^2 + m)
\end{split}
\end{equation}

Most recent metric learning loss functions are extensions of the triplet loss
\cite{DBLP:conf/cvpr/SchroffKP15,DBLP:conf/cvpr/SongXJS16,DBLP:conf/nips/Sohn16,DBLP:conf/cvpr/SongJR017,DBLP:conf/iccv/WangZWLL17,DDBLP:conf/cvpr/Wand2019,DBLP:conf/cvpr/WangHKHGR19,DBLP:conf/icml/LawUZ17,DBLP:conf/eccv/GeHDS18,DBLP:conf/iccv/ManmathaWSK17}, often combined with some type if intelligent sampling. Consequently, they are not based on a classification loss function, e.g., cross-entropy, but rather a loss function based on embedding distances. 
The rationale behind it is that what matters for a classification network is that the output is correct, which does not necessarily mean that the embeddings of samples belonging to the same class are similar. 
Since each sample is classified independently, it is entirely possible that two images of the same class have two distant embeddings that both allow for a correct classification. 
We argue that a classification loss can still be used for deep metric learning if the decisions do not happen independently for each sample, but rather jointly for a whole {\it group}, i.e., the set of images of the same class in a mini-batch. In this way, the method pushes for images belonging to the same class to have similar embeddings.
%

Towards this end, we propose {\it Group Loss} \cite{DBLP:conf/eccv/EleziVTPL20}, an iterative procedure that uses the global information of the mini-batch to refine the local information provided by the softmax layer of a neural network.
This iterative procedure categorizes samples into different \textit{groups}, and enforces consistent labelling among the samples of a group.
While softmax cross-entropy loss judges each sample in isolation, the Group Loss allows us to judge the overall class separation for {\it all} samples.
%
In section \ref{refine}, we show the differences between the softmax cross-entropy loss and Group Loss and highlight the mathematical properties of our new loss.

\subsection{Overview of Group Loss}\label{sec:gl_overview}

Given a mini-batch $\bm{B}$ consisting of $n$ images, consider the problem of assigning a class label $\bm{\lambda} \in \bm{\Lambda} = \{1, \dots, m\}$ to each image in $\bm{B}$. 
In the remainder of the manuscript, $\bm{X}=(\bm{x}_{i\bm{\lambda}})$ represents a $n \times m$ (non-negative) matrix of image-label soft assignments. In other words, each row of $\bm{X}$ represents a probability distribution over the label set $\bm{\Lambda}$ ($\sum_{\bm{\lambda}} x_{i\bm{\lambda}}=1 \mbox{ for all } i=1\dots n$). 

Our model consists of the following steps (see also Fig. \ref{fig:pipeline} and Algorithm \ref{algo}):

\begin{enumerate}
    \item \textbf{Initialization}: Initialize $\bm{X}$, the image-label assignment using the softmax outputs of the neural network. Compute the $n \times n$ pairwise similarity matrix $\bm{W}$ using the neural network embedding.
    \item \textbf{Refinement}: Iteratively, refine $\bm{X}$ considering the similarities between all the mini-batch images, as encoded in $\bm{W}$, as well as their labeling preferences.
    \item \textbf{Loss computation}: Compute the cross-entropy loss of the refined probabilities and update the weights of the neural network using backpropagation.
\end{enumerate}


\subsection{Initialization}

\textbf{Image-label assignment matrix.} The initial assignment matrix denoted $\bm{X}(0)$, comes from the softmax output of the neural network. 
We can replace some of the initial assignments in matrix $\bm{X}$ with one-hot labelings of those samples. We call these randomly chosen samples {\it anchors}, as their assignments do not change during the iterative refine process and consequently do not directly affect the loss function. However, by using their correct label instead of the predicted label (coming from the softmax output of the CNN), they guide the remaining samples towards their correct label. 

\textbf{Similarity matrix.} A measure of similarity is computed among all pairs of embeddings (computed via a CNN) in $\bm{B}$ to generate a similarity matrix $\bm{W} \in \mathbf{R}^{n \times n}$. 
In this work, we compute the similarity measure using the Pearson's correlation coefficient \cite{DBLP:journals/rsl/Pearson95}:
\begin{equation}\label{eq:pearson}
    \omega(i, j) = \frac{\mathrm{Cov}[\phi(\bm{I}_i), \phi(\bm{I}_j)]}{\sqrt{\mathrm{Var}[\phi(\bm{I}_i)]\mathrm{Var}[\phi(\bm{I}_j)]}} 
\end{equation} 
for $i\neq j$, and set $\mathcal{\omega}(i, i)$ to $0$.
The choice of this measure over other options such as cosine layer, Gaussian kernels, or learned similarities, is motivated by the observation that the correlation coefficient uses data standardization, thus providing invariance to scaling and translation -- unlike the cosine similarity, which is invariant to scaling only -- and it does not require additional hyperparameters, unlike Gaussian kernels \cite{DBLP:conf/icpr/EleziTVP18}. 
The fact that a measure of the linear relationship among features provides a good similarity measure can be explained by the fact that the computed features are actually a highly non-linear function of the inputs. Thus, the linear correlation among the embeddings actually captures a non-linear relationship among the original images.

\subsection{Refinement}
\label{refine}
In this core step of the proposed algorithm, the initial assignment matrix $\bm{X}(0)$ is refined in an iterative manner, taking into account the similarity information provided by matrix $\bm{W}$. $\bm{X}$ is updated in accordance with the \emph{smoothness assumption}, which prescribes that similar objects should share the same label.

To this end, let us define the \emph{support} matrix $\bm{\Pi}=(\bm{\pi}_{i\bm{\lambda}}) \in R^{n \times m}$ as
\begin{equation}
    \bm{\Pi} = \bm{WX} \label{eqn:pi}
\end{equation}
whose $(i, \lambda)$-component
\begin{equation}
    \pi_{i\lambda} = \sum_{j=1}^{n}w_{ij}x_{j\lambda}
\end{equation}
represents the \emph{support} that the current mini-batch gives to the hypothesis that the $i$-th image in $\bm{B}$ belongs to class $\lambda$. Intuitively, in obedience to the smoothness principle, $\pi_{i\lambda}$  is expected to be high if images similar to $i$ are likely to belong to class $\lambda$.

\begin{figure}[t]
    \centering
    \includegraphics[width=0.49\textwidth, trim={0.7cm 0.6cm 7cm 0cm},clip]{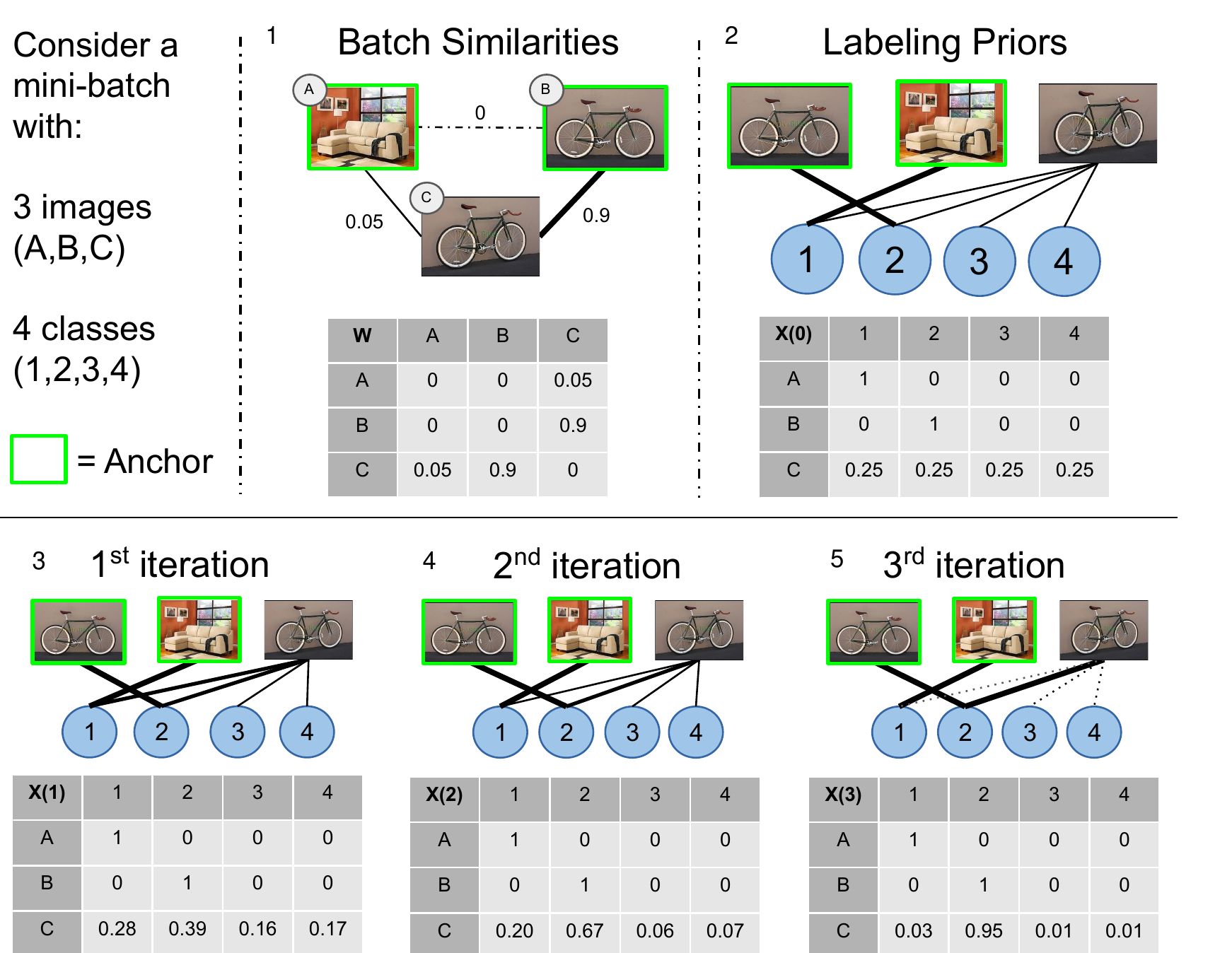}
    \caption{
    A  toy example of  the  refinement  procedure, where the goal is to classify sample C based on the similarity with samples A and B. (1) The Affinity matrix used to update the soft assignments. (2) The initial labeling of the matrix. (3-4) The process iteratively refines the soft assignment of the unlabeled sample C. (5) At the end, sample C gets the same label of A, (A, C) being more similar than (B, C).}
  \label{fig:gtg_prior}
\end{figure}

Given the initial assignment matrix $X(0)$, our algorithm refines it using the following update rule:
\begin{equation}
x_{i\lambda}(t+1) = \frac{x_{i\lambda}(t)\pi_{i\lambda}(t)}{\sum_{\mu=1}^m{x_{i\mu}(t)\pi_{i\mu}(t)}}
\label{eqn:rd}
\end{equation}
where the denominator represents a normalization factor which guarantees that the rows of the updated matrix sum up to one. 
This is known as multi-population replicator dynamics in evolutionary game theory \cite{weibull1997evolutionary} and is equivalent to nonlinear relaxation labeling processes \cite{RosHumZuc76,DBLP:journals/jmiv/Pelillo97}.

In matrix notation, the update rule (\ref{eqn:rd}) can be written as:
\begin{equation}
\bm{X}(t+1) = \bm{Q}^{-1}(t) \left[ \bm{X}(t) \odot \bm{\Pi}(t) \right] \label{eq:rd_mat}
\end{equation}
where 
\begin{equation}
\bm{Q}(t) = \mbox{diag}(\left[\bm{X}(t) \odot \bm{\Pi}(t)\right] \mathds{1} )
\end{equation}
and $\mathds{1}$ is the all-one $m$-dimensional vector.
$\bm{\Pi}(t) = \bm{WX}(t)$ as defined in (\ref{eqn:pi}), and $\odot$ denotes the Hadamard (element-wise) matrix product.
In other words, the diagonal elements of $\bm{Q}(t)$ represent the normalization factors in (\ref{eqn:rd}), which can also be interpreted as the average support that object $i$ obtains from the current mini-batch at iteration $t$.
Intuitively, the motivation behind our update rule is that at each step of the refinement process, for each image $i$, a label $\lambda$ will increase its probability $x_{i\lambda}$ if and only if its support $\pi_{i\lambda}$ is higher than the average support among all the competing label hypothesis $\bm{Q}_{ii}$. This can be motivated by a Darwinian survival-of-the-fittest selection principle, see e.g. \cite{weibull1997evolutionary}.

Thanks to the Baum-Eagon inequality \cite{DBLP:journals/jmiv/Pelillo97}, which has recently gained a renewed interest as an instance of Multiplicative Weights Update (MWU) rule \cite{NIPS2017_e93028bd, pmlr-v97-panageas19a}, it is easy to show that the dynamical system defined by (\ref{eqn:rd}) has very nice convergence properties. In particular, it strictly increases at each step the following functional:

\begin{equation}
\mathcal{F}(\bm{X}) = \sum_{i = 1}^{n}\sum_{j = 1}^{n}\sum_{\bm{\lambda} = 1}^{m} \bm{w}_{ij}  \bm{x}_{i\bm{\lambda}} \bm{x}_{j\bm{\lambda}} \label{eqn:avgconsistency}
\end{equation}
provided that the weights $\bm{w}_{ij}$ are non-negative. The functional (\ref{eqn:avgconsistency}) represents a measure of "consistency" \cite{hummelzucker} of the assignment matrix $\bm{X}$, in accordance to the smoothness assumption ($\mathcal{F}$ rewards assignments where highly similar objects are likely to be assigned the same label).
In other words:
\begin{equation}
\mathcal{F}(\bm{X}(t+1)) \geq \mathcal{F}(\bm{X}(t))
\label{eq8}
\end{equation}
with equality if and only if $\bm{X}(t)$ is a stationary point.
Hence, our update rule (\ref{eqn:rd}) is, in fact, an algorithm for maximizing the functional $\mathcal{F}$ over the space of row-stochastic matrices. Baum and Eagon \cite{baumeagon} further prove a more general result for non-homogeneous polynomials, with non-negative coefficients by proving that such mappings increase \emph{homotopically}, which in our case means that for $0 \le \eta \le 1$:
\begin{equation}
    \mathcal{F}(\eta\bm{X}(t+1) + (1 - \eta)\bm{X}(t)) \geq \mathcal{F}((\bm{X}(t))
\label{eq9}
\end{equation}
Note, that this contrasts with classical gradient methods, for which an increase in the objective function is guaranteed only when infinitesimal steps are taken, and determining the optimal step size entails computing higher-order derivatives. Here, instead, the step size is implicit and yet, at each step, the value of the functional increases.


\subsubsection{Dealing with negative similarities}
Equation (6) assumes that the matrix of similarity is non-negative.
However, for similarity computation, we use a correlation metric (see Equation (3)) which produces values in the range $[-1, 1]$.
In similar situations, different authors propose different methods to deal with the negative outputs.
The most common approach is to shift the matrix of similarity towards the positive regime by subtracting the biggest negative value from every entry in the matrix \cite{DBLP:journals/neco/ErdemP12}.
Nonetheless, this shift has a side effect: if a sample of class $k_1$ has very low similarities to the elements of a large group of samples of class $k_2$, these similarity values (which after being shifted are all positive) will be summed up. If the cardinality of class $k_2$ is very large, then summing up all these small values lead to a large value, and consequently affect the solution of the algorithm.
What we want instead, is to ignore these negative similarities, hence we propose {\it clamping}. More concretely, we use a \textit{ReLU} activation function over the output of Equation (3).
%

\subsection{Loss computation}
Once the labeling assignments converge (or in practice, a maximum number of iterations is reached), we apply the cross-entropy loss to quantify the classification error and backpropagate the gradients. Recall, the refinement procedure is optimized via \textit{replicator dynamics}, as shown in the previous section. By studying Equation (\ref{eq:rd_mat}), it is straightforward to see that it is composed of fully differentiable operations (matrix-vector and scalar products), and so it can be easily integrated within backpropagation. 
Although the refining procedure has no parameters to be learned, its gradients can be backpropagated to the previous layers of the neural network, producing, in turn, better embeddings for similarity computation.

\subsection{Summary of Group Loss} 
In this section, we proposed the Group Loss function for deep metric learning. 
During training, the Group Loss works by grouping together similar samples based on both the similarity between the samples in the mini-batch and the local information of the samples. The similarity between samples is computed by the correlation between the embeddings obtained from a CNN, while the local information is computed with a softmax layer on the same CNN embeddings. Using an iterative procedure, we combine both sources of information and effectively bring together embeddings of samples that belong to the same class.

During inference, we simply forward pass the images through the neural network
to compute their embeddings, which are directly used for image retrieval within a nearest neighbor search scheme.
The iterative procedure is not used during inference, thus making the feature extraction as fast as that of any other competing method. 

\begin{algorithm}[t] 
\caption{The Group Loss}
\label{algo}
\DontPrintSemicolon 
  \KwInput{input : Set of pre-processed images in the mini-batch $\bm{B}$, set of labels $\bm{y}$, 
  neural network $\phi$ with learnable parameters $\theta$, similarity function $\bm{\omega}$, number of iterations $T$}  
  
  Compute feature embeddings $\phi(\bm{B}, \theta)$ via the forward pass

  Compute the similarity matrix $\bm{W} = [\bm{\omega}(i, j)]_{ij}$
  
  Initialize the matrix of priors $\bm{X}(0)$ from the softmax layer
  
  \For{t = 0, \dots, T-1}
  {
    $\bm{Q}(t) = \mbox{diag}(\left[\bm{X}(t) \odot \bm{\Pi}(t)\right] \bm{1} )$\\
    $\bm{X}(t+1) = \bm{Q}^{-1}(t) \left[ \bm{X}(t) \odot \bm{\Pi}(t) \right]$
  }
      
  Compute the cross-entropy $\mathcal{J}(\bm{X}(T), \bm{y})$
  
  Compute the derivatives $\partial \mathcal{J} / \partial \theta$ via backpropagation, and update the weights $\theta$
\end{algorithm}

\section{Group Loss++: inference strategies}
\label{sec:inference_tricks}
In this section, we study the geometry of the embedding space that is built by the neural network trained with Group Loss. 
For each observation we make, we propose an inference strategy to further exploit such embedding geometry, increasing the results of retrieval without retraining.
%

\label{sec:inference_tricks}
\subsection{$\beta$-Normalization}
The embedding vector of an image is characterized by its length and direction. The softmax loss is mostly determined by the direction of an embedding pointing in the same direction as the weights of the correct class. Due to this property, the Group Loss embeddings are mainly trained to have a high inter-class and low intra-class cosine similarity. 
No further constraints are put on the embedding space by the loss. This results in embeddings that have a low loss only because they have high cosine similarity (point in the same direction) with the weights of the correct class despite having very different lengths.

By analyzing the embedding space, we observe that there is a large variance in the length of the embedding vectors, while the mean embedding length per class is very similar for all the classes. This means that feature vectors in the embedding space are often close due to being of similar length and not because they point in similar directions. Most state-of-the-art metric learning methods, including the original formulation of Group Loss \cite{DBLP:conf/eccv/EleziVTPL20}, address this issue by applying Euclidean normalization. Using Euclidean normalization on inference assumes that all of the discriminatory information of the embedding vector is stored in its direction. However, by doing so, all the information that the softmax layer stores in the length of the vector is lost. Since longer vectors get classified with more confidence by the softmax loss, the network learned to store discriminatory information in the length of the embeddings too.


In this work, we propose $\beta$-Normalization, a weighted normalization in which the information stored in the length of the vector is included but controlled by a parameter $\beta$. 
Using the embedding vector $\phi(\bm{I}_i)$, to 
control the degree of vector length information to the final embedding by modifying the scalar $\beta$ in the following equation:


\begin{equation}
\phi(\bm{I}_i)_{\beta}=\frac{\phi(\bm{I}_i)}{||\phi(\bm{I}_i)||_2}+\beta \phi(\bm{I}_i).
\end{equation}

One can quickly observe that 0-Normalization is equivalent to euclidean normalization and 1-Normalization barely has any influence on the original embeddings since we observed that the euclidean length of the embedding vectors is orders of magnitude larger than $1$. Unlike euclidean normalization, $\beta$-normalization enables us to make a compromise on the informativeness of length and direction which, in practice, leads to better performance.

\subsection{Mixed pooling}
Most metric learning works use Inception \cite{DBLP:conf/cvpr/SzegedyLJSRAEVR15} with batchnorm \cite{DBLP:conf/icml/IoffeS15}, ResNets \cite{7780459} or DenseNets \cite{DBLP:conf/cvpr/HuangLMW17}. All these networks use average pooling to transform the final feature volume into the embedding vector. 
Any pooling operation used to aggregate features will incur a loss of information. In the case of average pooling, the aggregation is performed averaging all the activations which nicely summarizes the global context of the feature map. Nonetheless, this also means that very large or very small values will be diluted in the average operation, potentially causing a loss of important information.
One can argue that we can simply replace average pooling with max pooling. However, max-pooling comes with its own issues as it assumes that most of the information is present in the maximum value of a feature map. 
We propose to use mixed pooling \cite{DBLP:conf/rskt/YuWCW14}, an operation that has been widely used in CNNs (e.g. person Re-ID), but with the exception of \cite{DBLP:journals/corr/abs-2004-01113} has not been used in the field of metric learning. We use both max and average pooling, combining them to to capture the global context of average pooling as well as the emphasis on the important information contained in large activation values that comes from max pooling.

\subsection{The effect of the final activation functions}
Most CNNs use the rectified linear unit (ReLU) as an activation function, which implies all the entries in the final feature volume before the final pooling contain positive values. Since max and average pooling only select and aggregate these positive values, the resulting embeddings are positive for every entry. 
From a geometric point of view, enforcing positive values heavily constrains the embedding space. For an $n$-dimensional embedding space, we only use $\frac{1}{2^{n}}$ of all the available space. To circumvent this issue and leverage the discriminatory information contained in negative activations, we argue that we should replace the final ReLU activation function with a leaky ReLU \cite{leaky_relu} on inference. This way we can control the influence that negative activations have on the final embedding by varying the slope in the leaky ReLU. Therefore possible discriminatory information contained in negative activation's can be translated into the embedding in a controlled fashion and is not lost.

\subsection{Re-ranking}
In person re-identification, a commonly used post-processing technique to make results more robust at test time is to perform k-reciprocal re-ranking \cite{DBLP:conf/cvpr/ZhongZCL17}. The intuition of it is to filter out false matches from the nearest neighbours (NN) of a given query image of class $y_i$. Furthermore, a gallery sample of class $y_i$ might be contained in the NN of some NN of a query image but not in the NN of query image itself. For example, if a query image is taken from the front, an image taken from diagonally behind might not be contained in the NN of the query itself but an image taken from the side might lie very close to both images in the embedding space. To also take such images into account, the NN of the query are extended.

To this end, we generate the reciprocal nearest neighbour of size $k_1$ ($k_1$-RNN) set of a given query image. This set only contains the gallery images from the query's $k_1$-nearest neighbour ($k_1$-NN) set that also contain the query image in their own $k_1$-NN set. Then, we expand the $k_1$-RNN set of the query image by $\frac{1}{2}k_1$-RNN sets of the gallery images that are contained in the query's $k_1$-RNN set and whose $\frac{1}{2}k_1$-RNN sets are sufficiently similar to the query's $k_1$-RNN sets, i.e., if their intersection is larger than some margin. To ensure that the expanded $k_1$-RNN set does not get too large, its size gets restricted to $k_2$.
Finally, we compute the Jaccard distance between the updated $k_1$-RNN sets of the query and gallery images. We use the weighted sum between the initial distance and the Jaccard distance with the weighting parameter $\lambda$ to rank the gallery images. 

While k-reciprocal re-ranking is omnipresent in person re-identification, and has shown impressive performance improvements, it was originally developed in the field of image retrieval \cite{k-reciprocal}. However, in the deep learning era, the vast majority of works do not utilize it.
In this work, we apply re-ranking in the task of image retrieval, showing that it consistently improves results, and thus should be a part of the retrieval toolbox. 

\subsection{Flip inference}
We flip every image and feed to the network both the original and the flipped image. We take the two resulting embeddings and average them. This makes the embedding invariant to flips in the input image. Introducing this invariance leads to superior embeddings because the discriminating features that were learned during training are included more optimally in the final embedding. In other words, if we had only seen birds from the left side in training and we encounter birds photographed from the right side during testing, the learned features that are strongly dependent on the side from which we look at the bird have little to no discriminatory power when Flip inference is not used. This intuition is further strengthened by the observation that for the Datasets where the images are fairly symmetric (In-Shop \cite{DBLP:conf/cvpr/LiuLQWT16}, Stanford Online Products \cite{DBLP:conf/cvpr/SongXJS16}) we gain much less than on datasets where the images are less symmetric (CUBS-200-2011 \cite{WahCUB_200_2011}, CARS196 \cite{KrauseStarkDengFei-Fei_3DRR2013}).

\section{Experimental Setup}

\subsection{Implementation details}
We use the PyTorch \cite{Paszke17} library for the implementation of Group Loss and Group Loss++. 

\textbf{Retrieval.} We use GoogleNet \cite{DBLP:conf/cvpr/SzegedyLJSRAEVR15} with batch-normalization \cite{DBLP:conf/icml/IoffeS15} as the backbone feature extraction. We pretrain the network on the \textit{ILSVRC 2012-CLS} dataset \cite{DBLP:journals/corr/RussakovskyDSKSMHKKBBF14}. For pre-processing, in order to get a fair comparison, we follow the implementation details of \cite{DBLP:conf/cvpr/SongJR017}. The inputs are resized to $256 \times 256$ pixels, and then randomly cropped to $227 \times 227$. Like other methods except for \cite{DBLP:conf/nips/Sohn16}, we use only a center crop during testing time. We train all networks for the classification task for $10$ epochs. We then train the network for the Group Loss task for $60$ epochs using RAdam optimizer \cite{DBLP:conf/iclr/LiuJHCLG020}. After $30$ epochs, we lower the learning rate by multiplying it by $0.1$. Following the works of \cite{DBLP:conf/icml/GuoPSW17,DBLP:conf/nips/BerthelotCGPOR19,DBLP:journals/corr/abs-1811-12649}, in order to calibrate the network, we use temperature scaling. We give the $\beta$ hyperparameter for normalization, and the hyperparameter for leaky relu in the appendix. We find the hyperparameters using random search \cite{DBLP:journals/jmlr/BergstraB12}. We use small mini-batches of size $30 - 100$. As sampling strategy, for each mini-batch, we first randomly sample a fixed number of classes, and then for each of the chosen classes, we sample a fixed number of samples. 

\textbf{Re-identification.} We use DenseNet161 \cite{DBLP:conf/cvpr/HuangLMW17} as backbone for the Group Loss. For fair comparison, we use the same pre-processing as in \cite{DBLP:conf/cvpr/0004GLL019}. To be specific, we resize all images to $256 \times 128$, pad the result with $10$ zero-valued pixels on each side and, finally, randomly crop the images again to $256 \times 128$. We train each network for $100$ epochs and reduce the learning rate by $10$ after $30$ and $60$ epochs using RAdam optimizer \cite{DBLP:journals/corr/abs-1908-03265}. We find all hyperparameters using random search \cite{DBLP:journals/jmlr/BergstraB12}. As commonly done in the field of person re-identification, we use random erasing \cite{Zhong2020RandomED}, where a random region of random size is erased from an image. This imitates occlusion, a common challenge in person re-identification, and therefore leads to a more robust performance of the model. We apply label smoothing as proposed in \cite{DBLP:conf/cvpr/KalayehBGKS18} to the auxiliary softmax cross-entropy loss right after the backbone CNN. Finally, inspired by \cite{DBLP:conf/cvpr/0004GLL019}, we add a batch normalization layer right before the final fully-connected layer that computes class probability scores. This is to ensure the feature vectors are balanced in all dimensions. Further, the features are normally distributed near the surface which, on the one hand, eases the convergence of the auxiliary softmax cross-entropy loss, and on the other hand, constrains the features belonging to the same person to be more compactly distributed \cite{DBLP:conf/cvpr/0004GLL019}. The latter also impacts the similarity computation of the Group Loss itself. As for the sampling strategy, for each mini-batch, we first sample a fixed number of classes with a low average distance between each other, and then for each class, we sample a fixed number of samples. 

\subsection{Benchmark datasets}
For image retrieval, we perform experiments on four publicly available datasets, evaluating our algorithm on both clustering and retrieval metrics. For training and testing, we follow the conventional splitting procedure \cite{DBLP:conf/cvpr/SongXJS16}.

\begin{itemize}
    \item \textbf{CUB-200-2011} \cite{WahCUB_200_2011} is a dataset containing $200$ species of birds with $11,788$ images, where the first $100$ species ($5,864$ images) are used for training and the remaining $100$ species ($5,924$ images) are used for testing.

    \item \textbf{Cars 196} \cite{KrauseStarkDengFei-Fei_3DRR2013} dataset is composed of $16,185$ images belonging to $196$ classes. We use the first $98$ classes ($8,054$ images) for training and the other $98$ classes ($8,131$ images) for testing.

    \item \textbf{Stanford Online Products} dataset \cite{DBLP:conf/cvpr/SongXJS16}, contains $22,634$ classes with $120,053$ product images in total, where $11,318$ classes ($59,551$ images) are used for training and the remaining $11,316$ classes ($60,502$ images) are used for testing.

    \item \textbf{In-Shop Clothes} dataset \cite{DBLP:conf/cvpr/LiuLQWT16} contains $7,982$ classes of clothing items and $4$ images on average in each class. $3,997$ classes are used for training, while the test set is split into a query set and a gallery set from $3,985$ classes.
\end{itemize}

\noindent For person re-identification we use the following datasets: 

\begin{itemize}

    \item \textbf{CUHK03 New Protocol} \cite{DBLP:conf/cvpr/LiZXW14}: The dataset comprises of 14,096 images of 14,097 identities, where each identity is captured from two different cameras. We use the new protocol \cite{DBLP:conf/cvpr/ZhongZCL17} where the identities are divided in 767 identities for training and 700 identities for testing.

    \item \textbf{Market1501} \cite{DBLP:conf/iccv/ZhengSTWWT15}: This dataset contains bounding boxes of 1,501 identities. Each identity is captured from at least two different and at most six different cameras. They are split into 751 identities containing 12,936 images and 750 identities containing 19,732 images for training and testing, respectively. All bounding boxes were automatically generated. 
\end{itemize}

\subsection{Evaluation metrics}
Based on the experimental protocol detailed above, we evaluate the retrieval performance and clustering quality on data from unseen classes of the $4$ aforementioned datasets. For the retrieval task, we calculate the percentage of the testing examples whose $K$ nearest neighbors contain at least one example of the same class. This quantity is also known as \textbf{Recall@K} \cite{DBLP:journals/pami/JegouDS11} and is the most used metric for image retrieval evaluation. The chosen value $K$ follows other works in the literature \cite{DBLP:conf/cvpr/SongXJS16}, and typically a larger $K$ is used for datasets that have a higher number of classes.
Similar to all other approaches, we perform clustering using K-means algorithm \cite{MacQueen} on the embedded features. Like in other works, we evaluate the clustering quality using the Normalized Mutual Information measure \textbf{(NMI)} \cite{DBLP:journals/corr/abs-1110-2515}. The choice of NMI measure is motivated by the fact that it is invariant to label permutation, a desirable property for cluster evaluation. 

For person re-identification, we use the two commonly used metrics: \textbf{CMC} that represents the percentage of query images whose top-$k$ ranked query images contain at least one image of the same identity (similar to Recall@K for retrieval); and \textbf{mAP} that represents the mean average precision of all query images.

\section{Results in image retrieval}

In this section, we will compare our baseline model, namely {\it Group Loss}, with the model with all inference tricks, {\it Group Loss++}. Furthermore, we compare our method with
In this section, we compare the Group Loss with state-of-the-art models on both image retrieval and clustering tasks.

\subsection{Comparison to SOTA}
\begin{table*}
\centering
    \resizebox{\textwidth}{!}{%
    \begin{tabular}{@{}l|c|ccccc|ccccc|cccc@{}}

\hline
\textbf{} & & \multicolumn{5}{c}{CUB-200-2011} & \multicolumn{5}{c}{CARS196} & \multicolumn{4}{c}{Stanford Online Products}\\ \hline
\textbf{Method} & \textbf{BB} & \textbf{R@1} & \textbf{R@2} & \textbf{R@4} & \textbf{R@8} & \textbf{NMI} &  \textbf{R@1} & \textbf{R@2} & \textbf{R@4} & \textbf{R@8} & \textbf{NMI} & \textbf{R@1} & \textbf{R@10} & \textbf{R@100} & \textbf{NMI} \\ \hline
Triplet \cite{DBLP:conf/cvpr/SchroffKP15} \textit{CVPR15} & G & 42.5 & 55 & 66.4 & 77.2 & 55.3 & 51.5 & 63.8 & 73.5 & 82.4 & 53.4 & 66.7 & 82.4 & 91.9 & 89.5 \\

Npairs \cite{DBLP:conf/nips/Sohn16} \textit{NeurIPS16} & G & 51.9 & 64.3 & 74.9 & 83.2 & 60.2 &  68.9 & 78.9 & 85.8 & 90.9 & 62.7 &  66.4 & 82.9 & 92.1 & 87.9  \\

Deep Spectral \cite{DBLP:conf/icml/LawUZ17} \textit{ICML17} & BNI & 53.2 & 66.1 & 76.7 & 85.2 & 59.2 &  73.1 & 82.2 & 89.0 & 93.0 & 64.3 &  67.6 & 83.7 & 93.3& 89.4 \\ 

Angular Loss \cite{DBLP:conf/iccv/WangZWLL17} \textit{ICCV17} & G & 54.7 & 66.3 & 76 & 83.9 & 61.1 &  71.4 & 81.4 & 87.5 & 92.1 & 63.2 &  70.9 & 85.0 & 93.5 & 88.6\\

Proxy-NCA \cite{DBLP:conf/iccv/Movshovitz-Attias17} \textit{ICCV17}& BNI & 49.2 & 61.9 & 67.9 & 72.4 & 59.5 &  73.2 & 82.4 & 86.4 & 88.7 & 64.9 &  73.7 & - & - & \textcolor{blue}{90.6} \\

Margin Loss \cite{DBLP:conf/iccv/ManmathaWSK17} \textit{ICCV17} & R50 & 63.6 & 74.4 & 83.1 & 90.0 & 69.0 &  79.6 & 86.5 & 91.9 & 95.1 & 69.1 &  72.7 & 86.2 & 93.8 & \textcolor{red}{90.7} \\

Hierarchical triplet \cite{DBLP:conf/eccv/GeHDS18} \textit{ECCV18} & BNI & 57.1 & 68.8 & 78.7 & 86.5 & - &  81.4 & 88.0 & 92.7 & 95.7 &  - & 74.8 & 88.3 & 94.8 & - \\

ABE  \cite{DBLP:conf/eccv/KimGCLK18} \textit{ECCV18} & G & 60.6 & 71.5 & 79.8 & 87.4 & - & 85.2 & 90.5 & 94.0 & 96.1 &  - & 76.3 & 88.4 &  94.8 & - \\

Normalized Softmax \cite{DBLP:journals/corr/abs-1811-12649} \textit{BMVC19} & R50 &  59.6 & 72.0 & 81.2 & 88.4 & 66.2 &  81.7 & 88.9 & 93.4 & 96.0 & 70.5 & 73.8 & 88.1 & 95 & 89.8 \\

RLL-H  \cite{DBLP:conf/cvpr/WangHKHGR19} \textit{CVPR19} & BNI & 57.4 & 69.7 & 79.2 & 86.9 & 63.6 & 74 & 83.6 & 90.1 & 94.1 &   65.4 &   76.1 & 89.1 & 95.4 & 89.7 \\

Multi-similarity  \cite{DDBLP:conf/cvpr/Wand2019} \textit{CVPR19} & BNI & 65.7 & 77.0 & 86.3 & 91.2 & - &  84.1 & 90.4 & 94.0 & 96.5 & - & 78.2 & 90.5 & 96.0 & - \\ 

Relational Knowledge \cite{DBLP:conf/cvpr/Park2019} \textit{CVPR19} & G & 61.4 & 73.0 & 81.9 & 89.0 & - &  82.3 & 89.8 & 94.2 & 96.6 & - & 75.1 & 88.3 & 95.2  & -\\ 

Divide and Conquer \cite{DDBLP:conf/cvpr/Sanakoyeu2019} \textit{CVPR19} & R50 & 65.9 & 76.6 & 84.4 & 90.6 & 69.6 & 84.6 & 90.7 & 94.1 & 96.5 &  70.3 &   75.9   & 88.4 & 94.9 &90.2 \\ 


SoftTriple Loss \cite{DBLP:journals/corr/abs-1909-05235} \textit{ICCV19}  & BNI & 65.4 & 76.4 & 84.5 & 90.4 & 69.3 & 84.5 & 90.7 & 94.5 & 96.9 & 70.1 & 78.3 & 90.3 & 95.9 & \textbf{92.0} \\

HORDE \cite{DBLP:journals/corr/abs-1908-02735} \textit{ICCV19} & BNI & 66.3 & 76.7 & 84.7 & 90.6 & -  & 83.9 & 90.3 & 94.1 & 96.3 & -  & \textbf{80.1} & 91.3 & \textcolor{blue}{96.2} & -\\

MIC \cite{DBLP:journals/corr/abs-1909-11574} \textit{ICCV19} & R50 & 66.1 & 76.8 & 85.6 & - & 69.7 & 82.6 & 89.1 & 93.2 & - & 68.4 & 77.2 & 89.4 & 95.6 & 90.0  \\

Easy triplet mining \cite{DBLP:conf/wacv/XuanSP20} \textit{WACV20} & R50 & 64.9 & 75.3 & 83.5 & - & - & 82.7 & 89.3 & 93.0 & - & - & 78.3 & 90.7 & \textcolor{red}{96.3} & - \\   

Proxy NCA++ \cite{DBLP:journals/corr/abs-2004-01113} \textit{ECCV20} & R50 & 66.3 & \textcolor{blue}{77.8} & \textbf{87.7} & \textcolor{blue}{91.3} & \textcolor{red}{71.3} & 84.9 & 90.6 & 94.9 & \textcolor{blue}{97.2} & \textcolor{blue}{71.5} & \textcolor{red}{79.8} & \textbf{91.4} & \textbf{96.4} & - \\

Proxy Anchor \cite{DBLP:conf/cvpr/KimKCK20} \textit{CVPR20} & BNI & \textcolor{red}{68.4} & \textcolor{red}{79.2} & \textcolor{blue}{86.8} & \textcolor{red}{91.6} & -  & \textcolor{red}{86.1} & \textcolor{blue}{91.7} & \textcolor{blue}{95.0} & \textcolor{red}{97.3} & -  & 79.1 & \textcolor{red}{90.8} & \textcolor{blue}{96.2} & - \\

Proxy-GML \cite{DBLP:journals/corr/abs-2010-13636} \textit{NeurIPS20} & BNI & \textcolor{blue}{66.6} & 77.6 & 86.4 & - & \textcolor{blue}{69.8} & 85.5 & \textcolor{red}{91.8} & \textcolor{red}{95.3} & - & \textcolor{red}{72.4} & 78.0 & \textcolor{blue}{90.6} & \textcolor{blue}{96.2} & 90.2 \\ \hline

Group Loss \cite{DBLP:conf/eccv/EleziVTPL20} \textit{ECCV20} & BNI & 65.5 & 77.0 & 85.0 & \textcolor{blue}{91.3} & 69.0 & 85.6 & 91.2 & 94.9 & 97.0 & 72.7 & 75.1 & 87.5 & 94.2 & 90.8 \\ 




Group Loss++ & BNI &\textbf{72.6} & \textbf{80.5} & \textcolor{red}{86.2} & \textbf{91.2} & \textbf{71.9} & \textbf{90.4} & \textbf{93.8} & \textbf{96.0} & \textbf{97.5} & \textbf{73.9} & \textcolor{blue}{79.2} & 90.1 & 95.8 & \textbf{92.0} \\ \hline

\end{tabular}}

\caption{Retrieval and Clustering performance on \textit{CUB-200-2011}, \textit{CARS196} and \textit{Stanford Online Products} datasets. Bold indicates best, red second best, and blue third best results. BB=backbone, G=GoogLeNet, BNI=BN-Inception, and R50=ResNet50.}

\label{tab:res_loss}

\end{table*}

\begin{table}[!ht]
\vspace{-0.7cm}
\centering
\resizebox{0.49\textwidth}{!}{%
\begin{tabular}{@{}l|c|cccc@{}}
\hline

\textbf{Method} & \textbf{BB} & \textbf{R@1} & \textbf{R@10} & \textbf{R@20} & \textbf{R@40} \\ \hline
FashionNet \cite{DBLP:conf/cvpr/LiuLQWT16} \textit{CVPR16} & V & 53.0 & 73.0 & 76.0 & 79.0 \\
A-BIER \cite{DBLP:journals/corr/abs-1801-04815} \textit{PAMI20}  & G & 83.1 & 95.1 & 96.9 & 97.8 \\
ABE \cite{DBLP:conf/eccv/KimGCLK18} \textit{ECCV18} & G & 87.3 & 96.7 & 97.9 & 98.5 \\
Multi-similarity  \cite{DDBLP:conf/cvpr/Wand2019} \textit{CVPR19} & BNI & 89.7 & \textcolor{red}{97.9} & \textcolor{red}{98.5} & 99.1 \\
Learning to Rank \cite{DBLP:conf/cvpr/Cakir0XKS19} & R50 & 90.9 & 97.7 & \textcolor{red}{98.5} & 98.9 \\
HORDE \cite{DBLP:journals/corr/abs-1908-02735} \textit{ICCV19} & BNI & 90.4 & 97.8 & 98.4 & 98.9 \\
MIC \cite{DBLP:journals/corr/abs-1909-11574} \textit{ICCV19} & R50 & 88.2 & 97.0 & 98.0 & 98.8 \\
Proxy NCA++ \cite{DBLP:journals/corr/abs-2004-01113} \textit{ECCV20} & R50 & \textcolor{blue}{90.4} & \textbf{98.1} & \textbf{98.8}  & \textbf{99.2} \\
Proxy Anchor \cite{DBLP:conf/cvpr/KimKCK20} \textit{CVPR20} & BNI & \textbf{91.5} & \textbf{98.1} & \textbf{98.8} & \textcolor{red}{99.1} \\ \hline
GroupLoss \cite{DBLP:conf/eccv/EleziVTPL20} \textit{ECCV20} & BNI & 86.8 & 96.4 & 97.5 & 98.4 \\ 
GroupLoss++ & BNI & \textcolor{red}{90.9} & \textcolor{blue}{97.6} & \textcolor{blue}{98.4} & \textcolor{blue}{98.9} \\ \hline

\end{tabular}}
\caption{Retrieval performance on \textit{In Shop Clothes}. Bold indicates best, red second best, and blue third best results. BB=backbone, G=GoogLeNet, BNI=BN-Inception, V=VGG16, and R50=ResNet50}
\vspace{0.1cm}
\label{tab:res_inshop}
\end{table}


In Tab. \ref{tab:res_loss}, we present the results of \textit{Group Loss++} and compare with the results of \textit{Group Loss} and those of the other approaches. On the \textit{CUB-200-2011} dataset, our \textit{Group Loss++} outperforms the other approaches by a large margin, with the second-best model , Proxy-Anchor \cite{DBLP:conf/cvpr/KimKCK20} reaching $4.2$ percentage points($pp$) lower absolute accuracy in Recall@1 metric. On NMI metric, \textit{Group Loss++} achieves a score of $71.9$ which is  $0.6pp$ better than previous SOTA, the Proxy-NCA++ \cite{DBLP:journals/corr/abs-2004-01113}. Compared to \textit{Group Loss} \cite{DBLP:conf/eccv/EleziVTPL20} we reach $7.1pp$ higher accuracy in Recall@1 and $2.9pp$ higher score in NMI metric. Similarly, on \textit{Cars 196}, \textit{Group Loss++} achieves the best results on Recall@1, with Proxy-Anchor \cite{DBLP:conf/cvpr/KimKCK20} coming second with a $4.3pp$ lower score in Recall@1. We improve compared to \textit{Group Loss} by $4.8pp$ in Recall@1. On the same dataset, we reach the highest results in NMI metric, a $1.5pp$ improvement over Proxy-GML method \cite{DBLP:journals/corr/abs-2010-13636}. On \textit{Stanford Online Products}, \textit{Group Loss++} reaches competitive results in Recall@1, $0.9pp$ worse than HORDE (which uses larger images), improving over \textit{Group Loss} by $4pp$. On the same dataset, when evaluated on the NMI score, our \textit{Group Loss++} outperforms all the other methods. Finally, on In-Shop Clothes, \textit{Group Loss++} reaches the second best results when evaluated in Recall@1, $0.6pp$ behind Proxy-Anchor, but improving over \textit{Group Loss} by $4.1pp$. 

\subsection{The effect of the refinement procedure}
\begin{figure}
  \vspace{-1cm}
  \centering
  \includegraphics[width=0.45\textwidth]{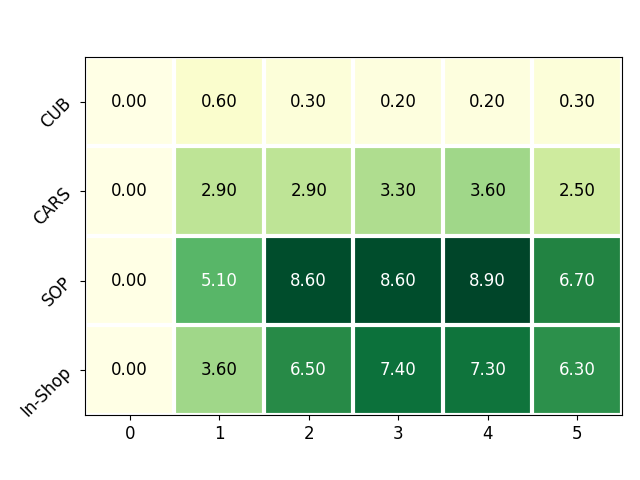}
  \vspace{-1cm}
  \caption{The effect of the number of refinement steps. $0$ represents the case where only cross-entropy is used, so there are no refinement steps.}
  \label{fig:refinement_steps}
\end{figure}

We validate our Group Loss method, by checking the effect of the refinement procedure, and its stability with respect to the number of refinement steps. We show the results in Fig. \ref{fig:refinement_steps}. In \textit{CUB-200-2011} dataset, we only see a marginal improvement compared to cross-entropy loss, with the best results being when we use only one refinement step, reaching $0.6pp$ better than when using cross-entropy. However, on the three other datasets, we reach significantly better results than cross-entropy, showing that the refinement procedure is the key to the success of Group Loss. More precisely, in \textit{Cars 196} dataset, we improve over cross-entropy by up to $3.6pp$ where we use four refinement steps; in \textit{Stanford Online Products}, we improve by $8.9pp$ where we use four refinement steps, and in \textit{In-Shop Clothes} we improve by up to $7.3-7.4pp$ where we use three, or four refinement steps. 

An interesting observation, is that the method performs best where there is a larger number of refinement steps (three or four), but the performance degrades where the number of refinement steps reaches five. The reasoning behind this is that for each image to be influenced not only by the images which are very similar to it (neighbours in context of nodes in the graph), but also by the images that are similar to its neighbors, it needs more than one refinement iteration. On the other hand, too many iterations, seem to make the performance worse, because every embedding starts becoming similar to each other. If too many iterations are done, every node starts effecting every other node, which is analogous to the problem of oversmoothing \cite{DBLP:conf/aaai/LiHW18} in graph neural networks \cite{DBLP:conf/iclr/KipfW17}.

\subsection{Inference ablation study}

\begin{table}
\centering
    \resizebox{0.49\textwidth}{!}{%
    \begin{tabular}{@{}l|cc|cc|cc|c@{}}

\hline
\textbf{} & \multicolumn{2}{c}{CUB-200-2011} & \multicolumn{2}{c}{CARS196} & \multicolumn{2}{c}{SOP} &
\multicolumn{1}{c}{In-Shop}\\ \hline
\textbf{Inference} &  \textbf{R@1} & \textbf{NMI} & \textbf{R@1} & \textbf{NMI} & \textbf{R@1} &  \textbf{NMI} & \textbf{R@1}  \\ \hline
GL & 65.5 & 69.0 & 85.6 & 72.7 & 75.1 & 90.8 & 86.8 \\
\hline

mixed & 67.5 & 69.5 & 88.2 & 72.9 & 78.1 & 91.2 & 89.1 \\
$\beta$ & 66.8 & 69.0 & 87.1 & 72.2 & 75.9 & 91.3 & 87.1 \\
L & 61.3 & 66.6 & 85.6 & 72.7 & 76.8 & 91.5 & 87.6 \\
R & 68.3 & 69.1 & 87.4 & 71.8 & 75.7 &  & 87.8 \\
F & 68.0 & 70.9 & 88.1 & 73.2 & 76.1 & 91.3 & 87.4 \\
\hline
GL++ & \textbf{72.6} & \textbf{71.9} & \textbf{90.4} & \textbf{73.9} & \textbf{79.2} & \textbf{92.0} & \textbf{90.9} \\
\hline


\end{tabular}}

\caption{The influence of different inference configurations on isolation. The last row represents the Results of Group Loss++. 
mixed=mixed pooling, $\beta$=$\beta$-normalization, L=LeakyReLU, R=re-ranking, F=Flip inference}

\label{tab:ablation_single}
\vspace{-0.5cm}
\end{table}

We now validate the inference choices described in the previous section. Our baseline is the Group Loss \cite{DBLP:conf/eccv/EleziVTPL20}, where we do not use any inference strategies, and that is denoted in Tab. \ref{tab:ablation_single}-\ref{tab:ablation_combined} as GL. 
In Tab. \ref{tab:ablation_single} we present the individual effect of each inference strategy. We show that replacing average pooling with mixed pooling, gives an improvement of $2-3$pp in the four datasets. We also see that beta-normalization in isolation gives a significant improvement, as do the re-ranking and inference flipping. On the other hand, replacing relu activation function with leaky-relu gives a moderate improvement in all datasets except CUB, where it reduces the performance by $4.2pp$.

\begin{table}
\vspace{-0.7cm}
\centering
    \resizebox{0.49\textwidth}{!}{%
    \begin{tabular}{@{}l|cc|cc|cc|c@{}}

\hline
\textbf{} & \multicolumn{2}{c}{CUB-200-2011} & \multicolumn{2}{c}{CARS196} & \multicolumn{2}{c}{SOP} &
\multicolumn{1}{c}{In-Shop}\\ \hline
\textbf{Inference} &  \textbf{R@1} & \textbf{NMI} & \textbf{R@1} & \textbf{NMI} & \textbf{R@1} &  \textbf{NMI} & \textbf{R@1}  \\ \hline
GL & 65.5 & 69.0 & 85.6 & 72.7 & 75.1 & 90.8 & 86.8 \\
\hline

mixed & 67.5 & 69.3 & 88.5 & 72.4 & 78.1 & 91.4 & 89.1 \\
mixed + $\beta$ & 68.0 & 70.2 & 88.5 & 72.8 & 78.0 & 91.8 & 88.9 \\
mixed + $\beta$ + L & 68.8 & 69.6 & 88.6 & 71.7 & 78.5 & \textbf{92.0} & 89.5 \\
mixed + $\beta$ + L + R & 70.2 & 70.2 & 89.3 & 72.1 & 78.9 & \textbf{92.0} & 90.4 \\
mixed + $\beta$ + L + R + F & \textbf{72.6} & \textbf{71.9} & \textbf{90.4} & \textbf{73.9} & \textbf{79.2} & \textbf{92.0} & \textbf{90.9} \\

\hline

\end{tabular}}

\caption{The influence of different inference configurations, cumulatively. The last row represents the Results of Group Loss++. 
mixed=mixed pooling, $\beta$=$\beta$-normalization, L=LeakyReLU, R=re-ranking, F=Flip inference}

\label{tab:ablation_combined}
\vspace{-0.5cm}
\end{table}

We present the cumulative effect of the inference strategies in Tab. \ref{tab:ablation_combined}. We see that adding each module comes with a significant improvement, showing that the strategies are complementary, hence we get the best results when we combine all of them.

\subsection{Implicit regularization and less overfitting.} 
\begin{figure}[t]
\centering
\begin{minipage}[t]{.24\textwidth}
  \centering
\includegraphics[width=\textwidth, trim={0cm 0cm 0cm 0cm},clip]{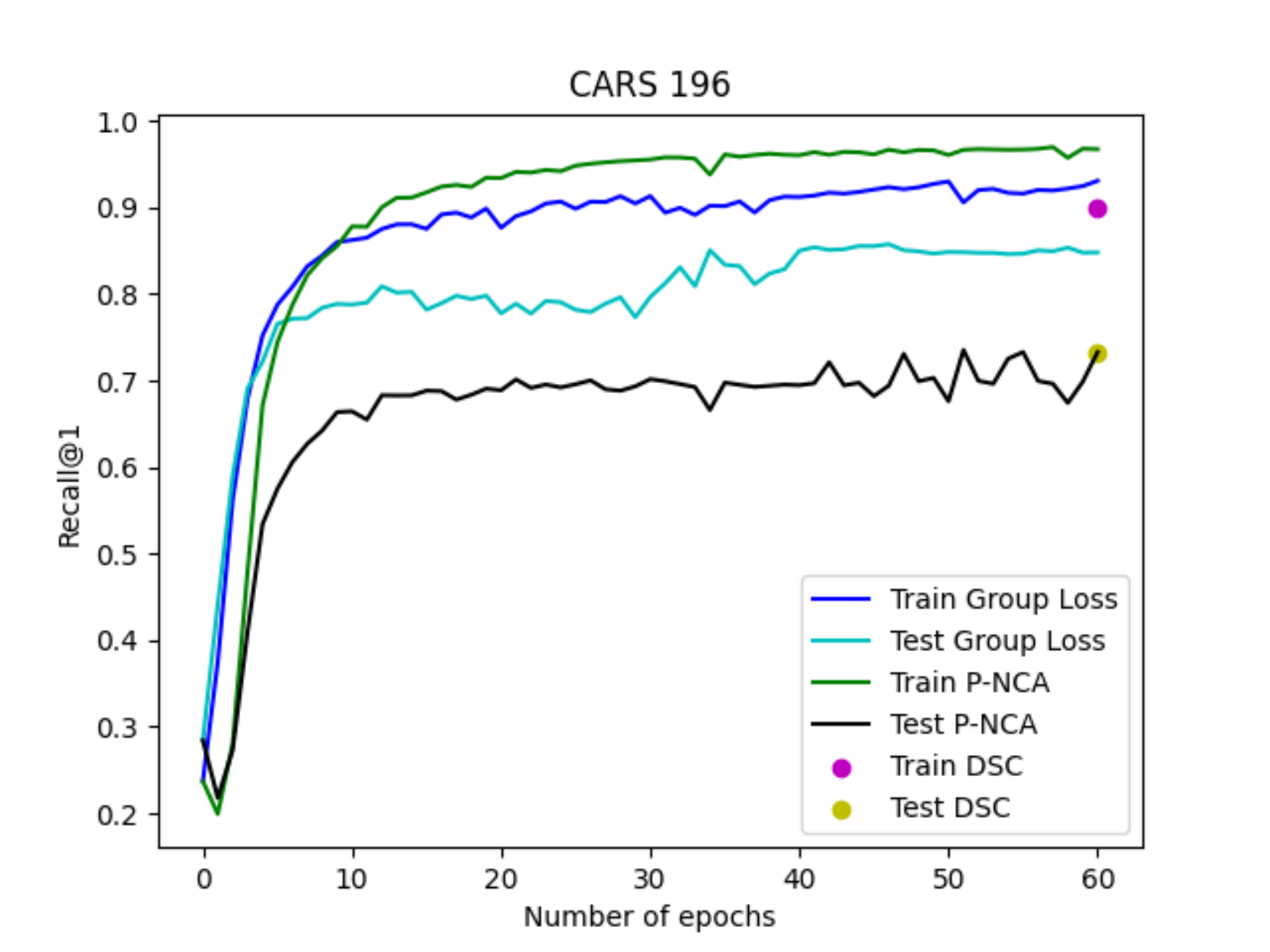} 
  \caption{Group Loss: Training vs testing Recall@1 curves on \textit{Cars 196} dataset.}
  \label{fig:regularization_cars}
\end{minipage}
\begin{minipage}[t]{.24\textwidth}
  \centering
\includegraphics[width=\textwidth, trim={0cm 0cm 0cm 0cm},clip]{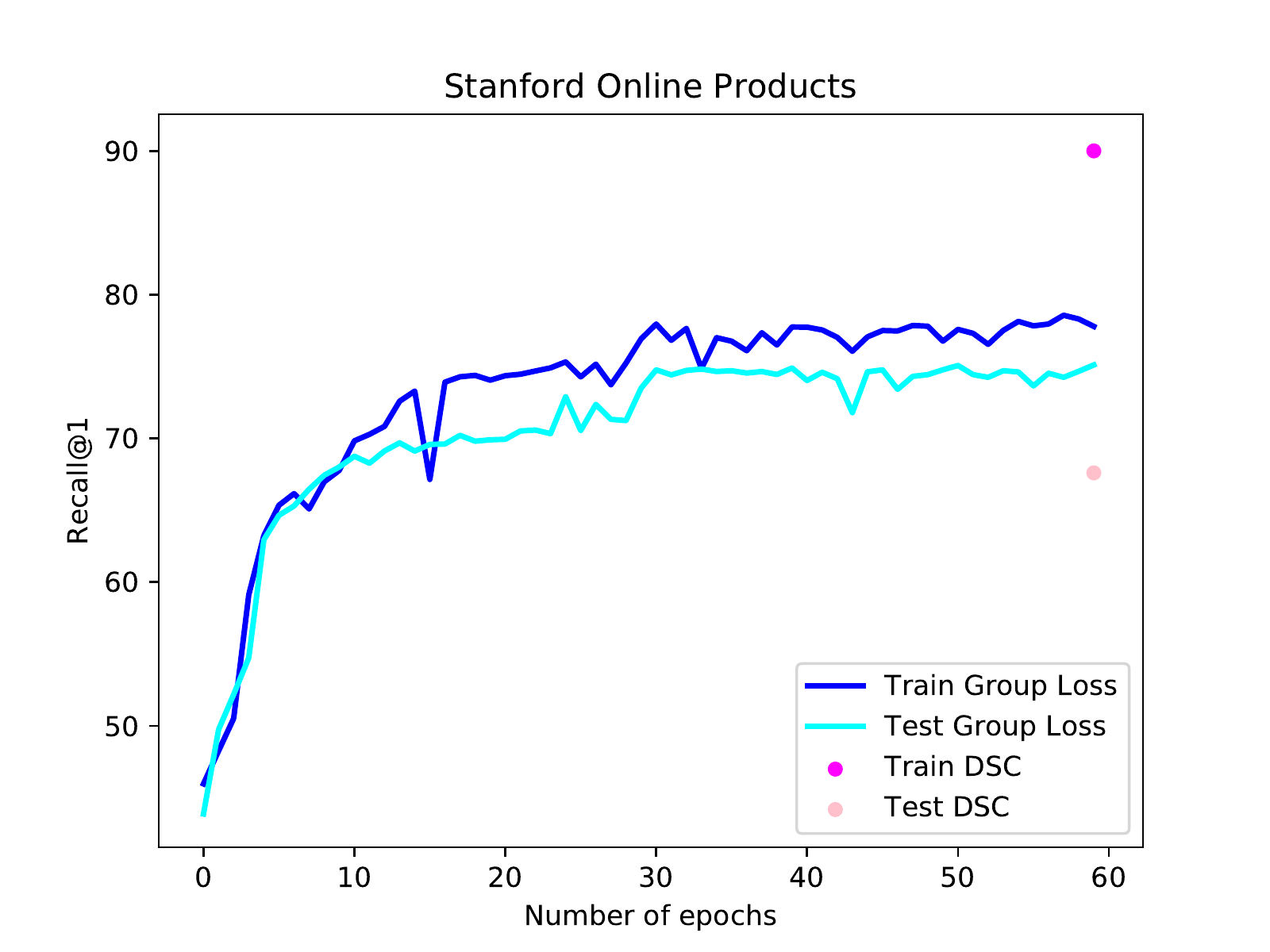} 
  \caption{Group Loss: Training vs testing Recall@1 curves on \textit{Stanford Online Products} dataset.}
  \label{fig:regularization_sop}
\end{minipage}
\end{figure}

In Figures \ref{fig:regularization_cars} and \ref{fig:regularization_sop}, we compare the results of training vs. testing on \textit{Cars 196} \cite{KrauseStarkDengFei-Fei_3DRR2013} and \textit{Stanford Online Products} \cite{DBLP:conf/cvpr/SongXJS16} datasets.
We see that the difference between Recall@1 at train and test time is small, especially on \textit{Stanford Online Products} dataset.
On \textit{Cars 196} the best results we get for the training set are circa $94\%$ in the Recall@1 measure, only 7.5 percentage points (pp) higher than what we reach with Group Loss in the  testing set (4pp higher than with the Group Loss++).
From the works we compared with, the only one which reports the results on the training set is 
\cite{DBLP:conf/icml/LawUZ17}. They reported results of over $90\%$ in 
all three datasets (for the training sets), much above the test set accuracy which lies at $73.1\%$ on \textit{Cars 196} and $67.6\%$ on \textit{Stanford Online Products} dataset. Unfortunately, they do not show the results after each epoch. \cite{DBLP:conf/wacv/VoH19} also provides results, but it uses a different network thus making the comparison impossible.

We further implement the P-NCA \cite{DBLP:conf/iccv/Movshovitz-Attias17} loss function and perform a similar experiment, in order to be able to compare training and test accuracy directly with our method. In Figure \ref{fig:regularization_cars}, we show the training and testing curves of P-NCA on the \textit{Cars 196} \cite{KrauseStarkDengFei-Fei_3DRR2013} dataset. We see that while in the training set, P-NCA reaches higher results than our method, in the testing set, our method outperforms P-NCA by almost $20pp$. Unfortunately, we were unable to reproduce the results of the paper \cite{DBLP:conf/iccv/Movshovitz-Attias17} on \textit{Stanford Online Products} dataset. Furthermore, even when we turn off $L2$-regularization, the generalization performance of our method does not drop at all.
Our intuition is that by taking into account the structure of the entire manifold of the dataset, our method introduces a form of regularization.
We can clearly see a smaller gap between training and test results when compared to competing methods, indicating less overfitting.

\section{Results in Person Re-Identification}
\label{sec:reid}
 Despite that the field of person re-identification is related to image retrieval, they have been studied separately. In this work, we provide a unified framework for image retrieval and person re-identification showing that our method works well for both problems. 

\subsection{Comparison to SOTA}
We compare the results of our method with state-of-the-art approaches on three re-identification datasets. In contrast to our approach, most of the approaches we compare to either take person parts into account \cite{DBLP:conf/cvpr/YangH0CHZ19,DBLP:conf/eccv/SunZYTW18,DBLP:conf/iccv/FangZRPH19,DBLP:conf/iccv/DaiCGZT19,DBLP:conf/iccv/QuanDWZY19,DBLP:conf/cvpr/MartinelFM19,DBLP:conf/eccv/SuhWTML18} or utilize features at different stages of the network \cite{DBLP:journals/corr/abs-1804-03864,DBLP:conf/cvpr/ZhangLZJ020,DBLP:conf/cvpr/ChenFZZSJY20,DBLP:conf/cvpr/ChenFZZSJY20,DBLP:conf/cvpr/Chen0LSW18} to get more fine-grained features. 

As can be seen in Table~\ref{tab:res_cuhk_np}, we report the best results in mAP score, reaching $1pp$ higher score than the previous SOTA on the CUHK03 dataset. In the rank-1 score, we reach the second-best results. We reach competitive results on the Market-1501 dataset, reaching $0.9pp$ worse results in rank-1 metric. Note, we reach these results without any specific adaption of the architecture on the person re-identification datasets, compared to methods like \cite{DBLP:conf/cvpr/ChenFZZSJY20} that uses three training stages.

\begin{table}
\centering
    \resizebox{0.32\textwidth}{!}{%
    \begin{tabular}{@{}l|c|cc@{}}

\hline
\textbf{} & & \multicolumn{2}{c}{CUHK03} \\ \hline
\textbf{Method}  & \textbf{BB} & \textbf{rank-1} & \textbf{mAP} \\ \hline
HA-Net \cite{DBLP:conf/cvpr/LiZG18} \textit{CVPR18} & HA-CNN & 41.7 & 38.6 \\
MLFN \cite{DBLP:conf/cvpr/ChangHX18} \textit{CVPR18} & MLFN & 52.8 & 47.8\\
CAMA \cite{DBLP:conf/cvpr/YangH0CHZ19} \textit{CVPR19} & R50 & 64.2 & 66.6  \\ 
MHN \cite{DBLP:conf/iccv/ChenDH19} \textit{ICCV19} & R50 & 71.1 & 65.4  \\ 
OSNet \cite{DBLP:conf/iccv/ZhouYCX19} \textit{ICCV19} & OSNet & 72.3 & 67.8  \\
Auto-ReID$^{\dagger}$ \cite{DBLP:conf/iccv/QuanDWZY19} \textit{ICCV19} & Auto-ReID & 73.3 & 69.9  \\
ISP \cite{DBLP:conf/eccv/ZhuGLTW20} \textit{ECCV20} & HRNet-W32 & 75.2 & 71.4 \\
BAT-net \cite{DBLP:conf/iccv/FangZRPH19} \textit{ICCV19} & BNI & 76.2 & 73.2 \\ 
BDB \cite{DBLP:conf/iccv/DaiCGZT19} \textit{ICCV19} & R50 & 76.4 & 73.5  \\  
RGA-SC \cite{DBLP:conf/cvpr/ZhangLZJ020} \textit{CVPR20} & R50 & 79.6 & 74.5  \\
PyrNet$^\dagger$ \cite{DBLP:conf/cvpr/MartinelFM19} \textit{CVPR19} & DN201 & \textcolor{blue}{80.8} & \textcolor{red}{82.7}\\
SCSN (3 stages) \cite{DBLP:conf/cvpr/ChenFZZSJY20} \textit{CVPR20} & R50 & \textbf{84.1} & \textcolor{blue}{80.2} \\
\hline

Group Loss & DN161 & 62.0 & 58.4 \\
Group Loss++ & DN161 & \textcolor{red}{82.5} & \textbf{83.7} \\ \hline


\end{tabular}}

\caption{Comparison to state-of-the-art results on the new test protocol and the detected bounding boxes of CUHK03 dataset \cite{DBLP:conf/cvpr/ZhongZCL17}.  Bold indicates best, red second best, and blue third best results.  BB=backbone, DN201=DenseNet-201, DN161=DenseNet-161, BNI=BN-Inception and R50=ResNet50. All other backbones are specifically designed to person re-identification. $\dagger$ means that either k-reciprocal re-ranking orsome other re-ranking approach is used.}
\vspace{-0.7cm}
\label{tab:res_cuhk_np}
\end{table}
\begin{table}
\centering
    \resizebox{0.32\textwidth}{!}{%
    \begin{tabular}{@{}l|c|cc@{}}

\hline
\textbf{} & \multicolumn{2}{c}{Market-1501}\\ \hline
\textbf{Method} & \textbf{BB} & \textbf{rank-1} & \textbf{mAP} \\ \hline
TriNet$^\dagger$ \cite{DBLP:journals/corr/HermansBL17} & R50& 91.8 & 87.2 \\ 
SGGNN$\dagger$ \cite{DBLP:conf/eccv/ShenLYCW18} \textit{ECCV18} & R50 & 92.3 & 82.2 \\ 
DGSRW$\dagger$  \cite{DBLP:conf/cvpr/ShenLXYCW18} \textit{CVPR18} & R50 & 92.7 & 82.5 \\ 
GCSL \cite{DBLP:conf/cvpr/Chen0LSW18} \textit{CVPR18} & R50& 93.5 & 81.6  \\ 
HA-Net \cite{DBLP:conf/cvpr/LiZG18} \textit{CVPR18} & HA-CNN & 93.8 & 82.2 \\
IANet \cite{DBLP:conf/cvpr/HouMCGSC19a} \textit{CVPR19} & R50 & 94.4 & 83.1\\ 
HSP$^\dagger$ \cite{DBLP:conf/cvpr/KalayehBGKS18} \textit{CVPR18} & BNI & 94.6 & 90.9 \\ 
CAMA \cite{DBLP:conf/cvpr/YangH0CHZ19} \textit{CVPR19} & R50 & 94.7 & 84.5 \\ 
OSNet \cite{DBLP:conf/iccv/ZhouYCX19} \textit{ICCV19}& OSNet & 94.8 & 84.9 \\ 
DG-Net \cite{DBLP:conf/cvpr/ZhengYY00K19} \textit{CVPR19}  & R50 & 94.8 & 86.0 \\ 
PCB \cite{DBLP:conf/eccv/SunZYTW18} \textit{ECCV18} & R50 & 95.1 & 91.9 \\ 
BAT-net \cite{DBLP:conf/iccv/FangZRPH19} \textit{ICCV19} & BNI & 95.1 & 87.4\\ 
MHN \cite{DBLP:conf/iccv/ChenDH19} \textit{ICCV19} & R50 & 95.1 & 85.0 \\ 
BDB \cite{DBLP:conf/iccv/DaiCGZT19} \textit{ICCV19} & R50 & 95.3 & 86.7 \\ 
ISP \cite{DBLP:conf/eccv/ZhuGLTW20} \textit{ECCV20} & HRNet-W32 & 95.3 & 88.6 \\
Auto-ReID$^\dagger$ \cite{DBLP:conf/iccv/QuanDWZY19} \textit{ICCV19} & AutoReID & 95.4 & \textbf{94.2} \\ 
BoT$^\dagger$ \cite{DBLP:conf/cvpr/0004GLL019} \textit{CVPR19} & R50 & 95.4 & \textbf{94.2} \\ 
DCDS$^\dagger$ \cite{DBLP:journals/corr/abs-1904-11397} \textit{ICCV19} & R101 & 95.4 & \textcolor{blue}{93.3} \\
ABDNet \cite{DBLP:conf/iccv/ChenDXYCYRW19} \textit{ICCV19} & R50 & \textcolor{blue}{95.6} & 88.3 \\
SCSN (3 stages) \cite{DBLP:conf/cvpr/ChenFZZSJY20} \textit{CVPR20} & R50 & \textcolor{red}{95.7} & 88.5 \\ 
PyrNet$^{\dagger}$ \cite{DBLP:conf/cvpr/MartinelFM19} \textit{CVPR19} & DN201 & \textbf{96.1} & \textcolor{red}{94.0} \\ 
RGA-SC \cite{DBLP:conf/cvpr/ZhangLZJ020} \textit{CVPR20} & R50 & \textbf{96.1} & 88.4 \\ 
\hline
Group Loss & DN161 & 93.7 & 78.3 \\
Group Loss++ & DN161 & 95.2 & 92.1 \\ \hline


\end{tabular}}
\caption{Comparison to state-of-the-art results on Market-1501 \cite{DBLP:conf/iccv/ZhengSTWWT15}. Bold indicates best, red second best, and blue third best results. BB=backbone, DN201=DenseNet-201, DN161=DenseNet-161, BNI=BN-Inception and R50=ResNet50. All other backbones are specifically designed to person re-identification. $\dagger$ means that either k-reciprocal re-ranking or some other re-ranking approach is used.}

\label{tab:res_market}
\vspace{-1.1cm}
\end{table}

\section{Discussion and Limitations}

\subsection{The connections with Graph Neural Networks}
Our method is connected to graph neural networks \cite{DBLP:journals/tnn/ScarselliGTHM09, DBLP:conf/iclr/KipfW17}, however, an advantage of our label-propagator is that it has guaranteed convergence properties, and eventually reaches a stationary point (Eq. \ref{eq8}, \ref{eq9}). Closest to this work, is our recently published method \cite{DBLP:journals/corr/abs-2102-07753} where the samples exchange feature preferences (instead of label preferences) based on a message-passing network \cite{DBLP:conf/icml/GilmerSRVD17} implemented via graph attention networks \cite{DBLP:conf/iclr/VelickovicCCRLB18} which can be seen as Transformers \cite{DBLP:conf/nips/VaswaniSPUJGKP17}. Considering that graph neural networks have been very successful in the field of semi-supervised learning, it would be interesting to extend our method towards semi-supervised tasks.
\vspace{-0.2cm}
\subsection{Dealing with Relative Labels}
The proposed method assumes that the dataset consists of classes with (absolute) labels and a set of samples in each class. This is the case for the datasets used in metric learning \cite{WahCUB_200_2011,KrauseStarkDengFei-Fei_3DRR2013,DBLP:conf/cvpr/SongXJS16}) technique evaluations. However, deep metric learning can be applied to more general problems where the absolute class label is not available but a relative label is available. For example, the data might be given as pairs that are similar or dissimilar. Similarly, the data may be given as triplets consisting of anchor (A), positive (P) and negative (N) images, such that A is semantically closer to P than N. For example, in \cite{DBLP:conf/iccv/WangG15} a triplet network has been used to learn good visual representation where only relative labels are used as self-supervision (two tracked patches from the same video form a "similar" pair and the patch in the first frame and a patch sampled from another random video forms a "dissimilar" pair). Similarly, in \cite{DBLP:conf/eccv/MisraZH16}, relative labels are used as self-supervision for learning good spatio-temporal representation. 

Our method, similar to other classification-based losses \cite{DBLP:journals/corr/abs-1811-12649,DBLP:conf/iccv/Movshovitz-Attias17,DBLP:journals/corr/abs-1909-05235} for deep metric learning, or triple loss improvements like Hierarchical Triplet Loss \cite{DBLP:conf/eccv/GeHDS18} assumes the presence of absolute labels. Unlike traditional losses \cite{bromley1994signature,DBLP:conf/cvpr/SchroffKP15}, in cases where only relative labels are present, all the mentioned methods do not work. However, in the presence of both relative and absolute labels, then in our method, we could use relative labels to initialize the matrix of similarities, potentially further improving the performance of networks trained with The Group Loss.

\subsection{Dealing with a Large Number of Classes}

In all classification-based methods, the number of outputs in the last layer linearly increases with the number of classes in the dataset. This can become a problem for learning on a dataset with a large number of classes (say $N>1000000$) where metric learning methods like pairwise/triplet losses/methods can still work. In our method, the similarity matrix is square on the number of samples per mini-batch, so its dimensions are the same regardless if there are $5$ or $5$ million classes. However, the matrix of probabilities is linear in the number of classes. If the number of classes grows, computing the softmax probabilities and the iterative process becomes indeed computationally expensive. An alternative, which reaches similar results, is to sparsify the matrix. Considering that in any mini-batch, we use only a small number of classes ($<10$), all the entries not belonging to these classes may be safely set to 0, followed by a normalization of the probability matrix. This allows both saving storage (e.g. using sparse tensors in PyTorch) and an efficient tensor-tensor multiplication. It also needs to be said, that in retrieval, the number of classes is typically not very large, and many other methods \cite{DBLP:journals/corr/abs-1811-12649,DBLP:conf/iccv/Movshovitz-Attias17,DBLP:journals/corr/abs-1909-05235,DBLP:conf/eccv/GeHDS18} face the same problem. However, in related fields, e.g., face recognition, there could be millions of classes (identities), in which case we can use the proposed solution.

\section{Conclusion}

In this work, we proposed the Group Loss, a new loss function for deep metric learning that goes beyond triplets. 
By considering the content of a mini-batch, it promotes embedding similarity across all samples of
the same class, while enforcing dissimilarity for elements of different classes.
This is achieved with a fully-differentiable layer that is used to train a CNN in an end-to-end fashion.
We then propose the Group Loss++, by modifying the original formulation to include inference strategies that significantly increase the performance of our model without any training changes. Finally, we show that our method can be used without changes for the task of person re-identification reaching competitive results. Our method reaches state-of-the-art results in several metric learning datasets.

\ifCLASSOPTIONcompsoc
  \section*{Acknowledgments}
\else
  \section*{Acknowledgment}
\fi

This research was partially funded by the Humboldt Foundation through the Sofja Kovalevskaja Award and a Humboldt Research Fellowship. 

\newpage
\section*{APPENDIX}
\section*{More implementation details}
\label{implementation_details}
We first pre-train all networks in the classification task for $10$ epochs. We then train our networks on all three datasets \cite{WahCUB_200_2011,KrauseStarkDengFei-Fei_3DRR2013,DBLP:conf/cvpr/SongXJS16} for $60$ epochs. During training, we use a simple learning rate scheduling in which we divide the learning rate by $10$ after the first $30$ epochs.

We find all training hyperparameters using random search \cite{DBLP:journals/jmlr/BergstraB12}. For the weight decay ($L2$-regularization) parameter, we search over the interval $[0.1, 10^{-16}]$, while for learning rate we search over the interval $[0.1, 10^{-5}]$, choosing $0.0002$ as the learning rate for all networks and all datasets.

We achieve the best results with a regularization parameter set to $10^{-6}$ for \textit{CUB-200-2011}, $10^{-7}$ for \textit{Cars 196} dataset, and $10^{-12}$ for \textit{Stanford Online Products} dataset. This further strengthens our intuition that the method is implicitly regularized and it does not require strong regularization. 

Apart from the training hyperparameters, we find the inference hyperparameters by manual search for each dataset individually as the hyperparameters are highly dependent on each other. The final values can be found in Tab. \ref{tab:inference_hypers}.

\section*{Robustness analysis with respect to hyperparameters}
\label{robustness_gl}

\begin{figure*}[ht!]
\centering
\begin{minipage}[t]{.49\textwidth}
  \centering
\includegraphics[width=\textwidth, trim={0cm 0cm 0cm 0cm},clip]{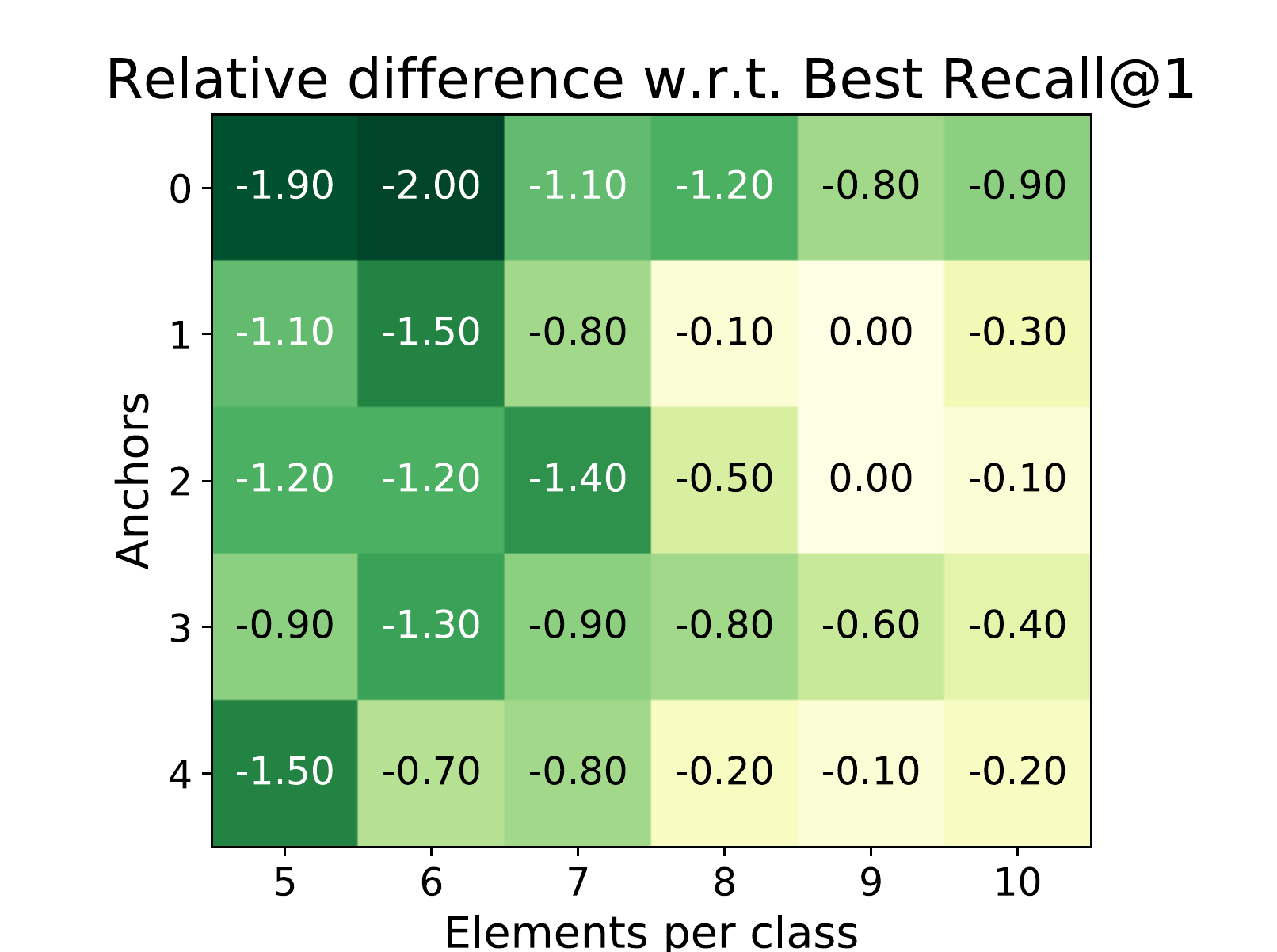} 
  \caption{The effect of the number of anchors and the number of samples per class in CUB-200-2011.}
  \label{fig:abl_labeledpoints}
\end{minipage}
\hfill %
\begin{minipage}[t]{.49\textwidth}
  \centering
\includegraphics[width=\textwidth, trim={0cm 0cm 0cm 0cm},clip]{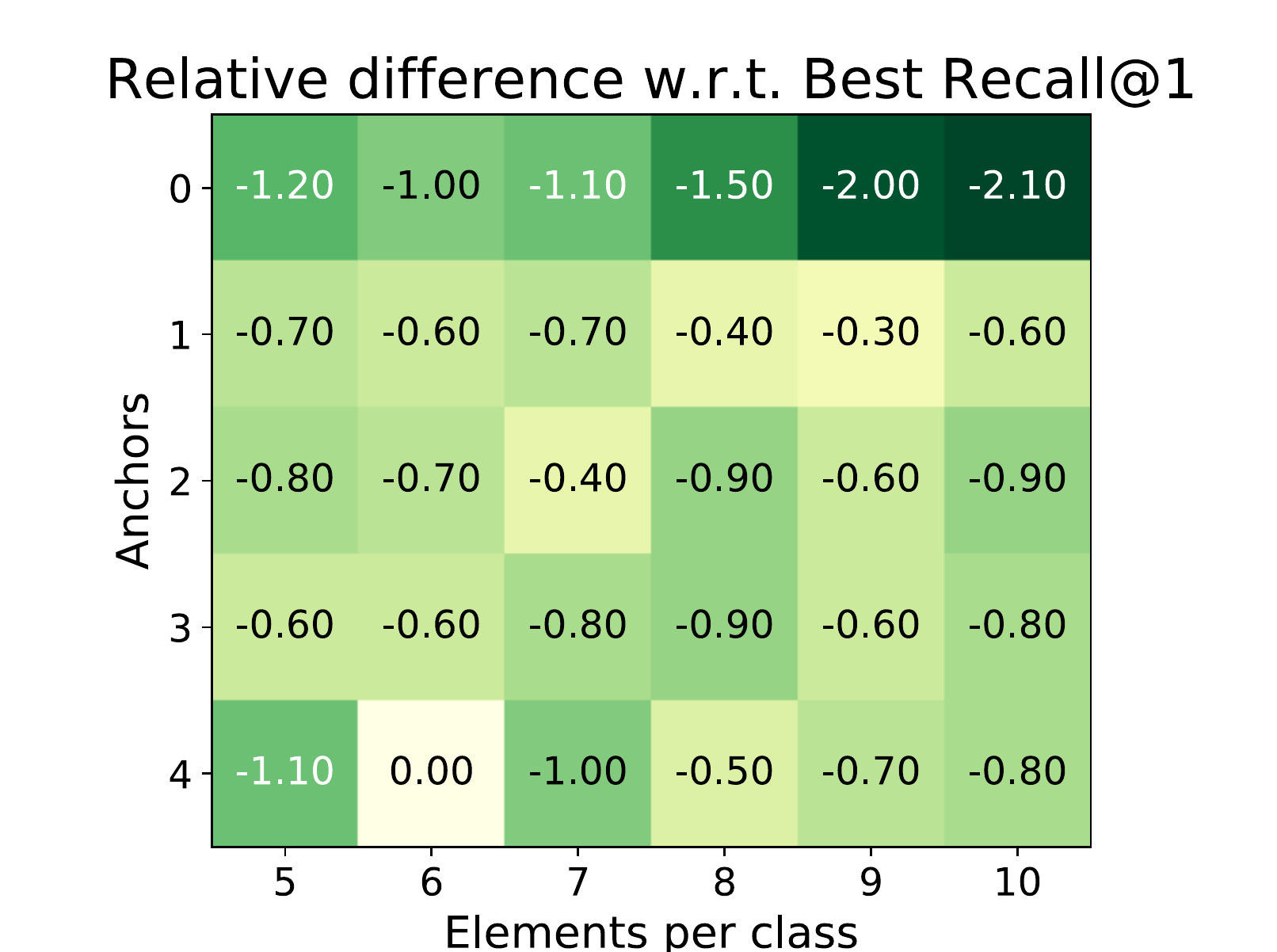} 
  \caption{The effect of the number of anchors and the number of samples per class in Cars 196.}
  \label{fig:abl_labeledpoints_cars}
\end{minipage}
\hfill %
\end{figure*}

\begin{figure*}[ht!]
\centering
\begin{minipage}[t]{.49\textwidth}
  \centering
\includegraphics[width=\textwidth, trim={0cm 0cm 0cm 0cm},clip]{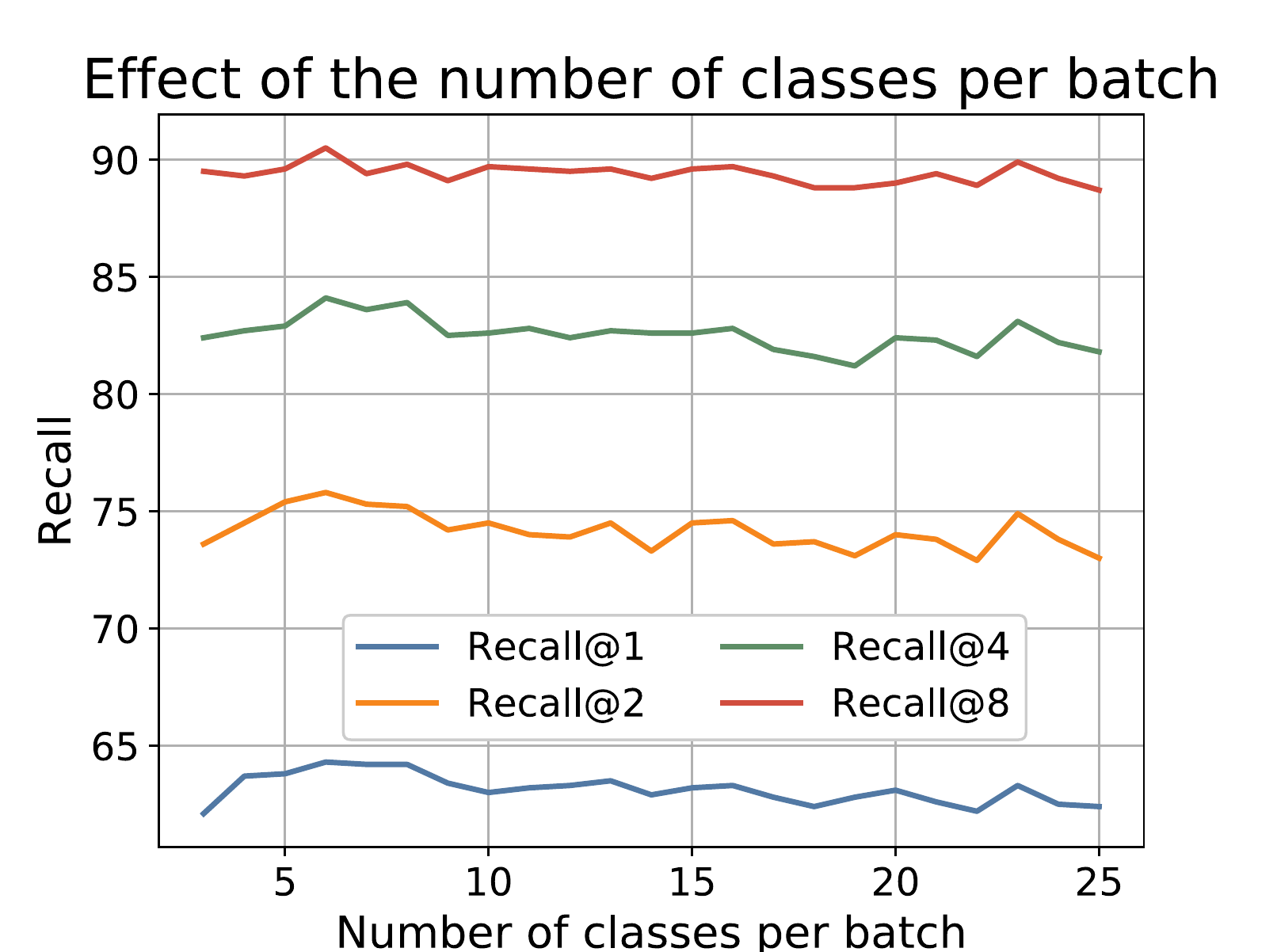}
  \caption{The effect of the number of classes per mini-batch.}
  \label{fig:abl_classes}
\end{minipage}
\hfill %
\begin{minipage}[t]{.49\textwidth}
  \centering
\includegraphics[width=\textwidth, trim={0cm 0cm 0cm 0cm},clip]{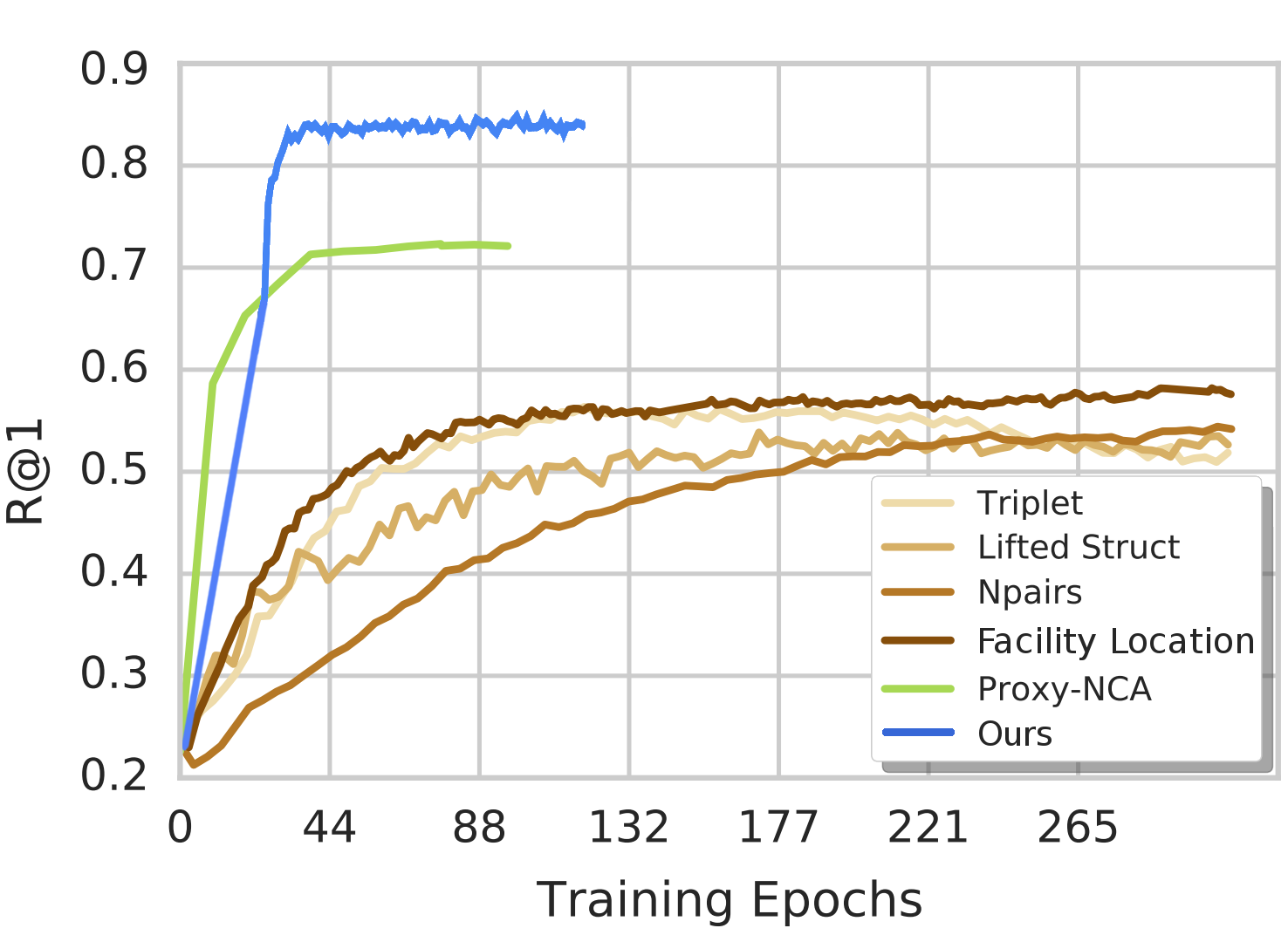} 
  \caption{Recall@1 as a function of training epochs on Cars 196 dataset. Figure adapted from \cite{DBLP:conf/iccv/Movshovitz-Attias17}.}
  \label{fig:abl_convergence}
\end{minipage}
\hfill %
\end{figure*}

\textbf{Number of anchors.} 
In Fig.~\ref{fig:abl_labeledpoints} and ~\ref{fig:abl_labeledpoints_cars}, we show the effect of the number of anchors with respect to the number of samples per class. We do the analysis on \textit{CUB-200-2011} \cite{WahCUB_200_2011} and \textit{Cars 196} \cite{KrauseStarkDengFei-Fei_3DRR2013} datasets. The results reported are the percentage point differences in terms of Recall@1 with respect to the best performing set of parameters. 
The number of anchors ranges from $0$ to $4$, while the number of samples per class varies from $5$ to $10$. 
It is worth noting that on \textit{CUB-200-2011} \cite{WahCUB_200_2011}, our best setting considers $1$ or $2$ anchors over $9$ samples. Moreover, even when we do not use any anchor, the difference in Recall@1 is no more than $2pp$.
Similarly, the method shows the same robustness as on \textit{Cars 196} \cite{KrauseStarkDengFei-Fei_3DRR2013}, with the best result being only $2.1$ percentage points better at the Recall@1 metric than the worst result.
Note that the results decrease mainly when we do not have any labeled sample, i.e., when we use zero anchors.

\textbf{Number of classes per mini-batch.}
In Fig.~\ref{fig:abl_classes}, we present the change in Recall@1 on the \textit{CUB-200-2011} \cite{WahCUB_200_2011} dataset if we increase the number of classes we sample at each iteration. The best results are reached when the number of classes is not too large. This is a welcome property, as we can train on small mini-batches, known to achieve better generalization performance \cite{DBLP:conf/iclr/KeskarMNST17}.

\begin{table}
\centering
    \resizebox{0.45\textwidth}{!}{%
    \begin{tabular}{@{}l|c|c|c|c|c|c@{}}

\hline
\textbf{dataset}  &  \textbf{$\beta$} & \textbf{LeakyReLU} & \textbf{$\alpha$} & \textbf{$\lambda$} & \textbf{$k_1$} & \textbf{$k_2$} \\ \hline
CUB-200-2011 & 0.004 & 0.75 & 0.5 & 0.85 & 50 & 20\\
Cars196 & 0.004 & 0.4 & 0.6 & 0.9 & 50 & 20\\
In-Shop & 0.002 & 1.0 & 0.9 & 0.7 & 4 & 2\\
Stanford Online Products & 0.0005 & 1.0 & 0.6 & 0.94 & 50 & 20 \\

\end{tabular}}

\caption{Hyperparameters of inference strategies for all four datasets.}

\label{tab:inference_hypers}

\end{table}

\section*{Sensitivity with Respect to Inference Hyperparameters}
\label{robustness_gl++}

In this section, we analyze the sensitivity of the performance of GL++ with respect to the inference hyperparameters. We do the analysis on the smaller \textit{CUB-200-2011} and \textit{Cars 196} datasets.

\begin{figure}

\begin{minipage}{\columnwidth}
\includegraphics[width=0.9\textwidth]{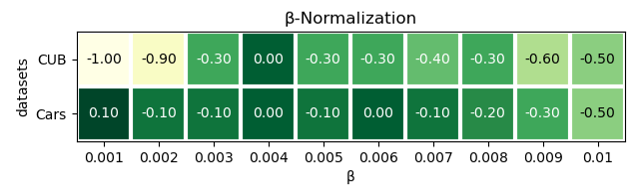}
\vspace{-0.4cm}
\caption{Sensitivity of R@1 with respect to $\beta$-normalization (equidistant).}
\vspace{+0.3cm}
\label{fig:beta}
\end{minipage}

\begin{minipage}{\columnwidth}
\includegraphics[width=0.9\textwidth]{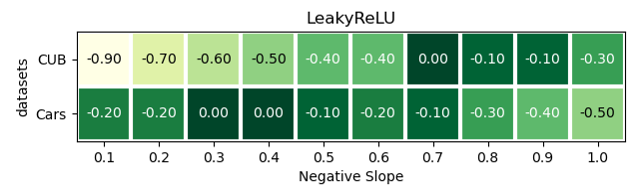}
\vspace{-0.4cm}
\caption{Sensitivity of R@1 with respect to negative slope of LeakyReLU.}
\vspace{+0.3cm}
\label{fig:leaky} 
\end{minipage}

\begin{minipage}{\columnwidth}
\includegraphics[width=0.9\textwidth]{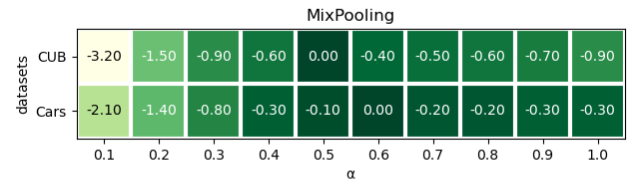}
\vspace{-0.4cm}
\caption{Sensitivity of R@1 with respect to different combinations of average and maximum pooling.}
\vspace{+0.3cm}
\label{fig:max}
\end{minipage}

\begin{minipage}{\columnwidth}
\includegraphics[width=0.9\textwidth]{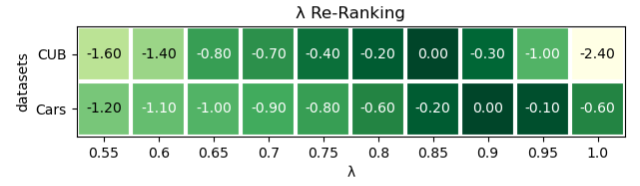}
\vspace{-0.4cm}
\caption{Sensitivity of R@1 with respect to different $\lambda-$values of re-ranking.}
\vspace{+0.3cm}
\label{fig:lamda_re_ranking}
\end{minipage}

\begin{minipage}{\columnwidth}
\includegraphics[width=0.9\textwidth]{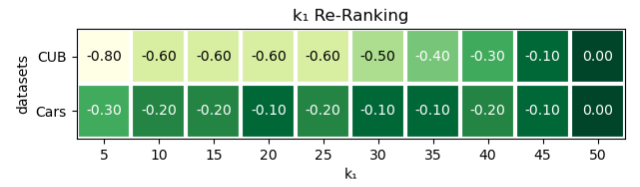}
\vspace{-0.4cm}
\caption{Sensitivity of R@1 with respect to different $k1-$values of re-ranking.}
\vspace{+0.3cm}
\label{fig:k1}
\end{minipage}

\begin{minipage}{\columnwidth}
\includegraphics[width=0.9\textwidth]{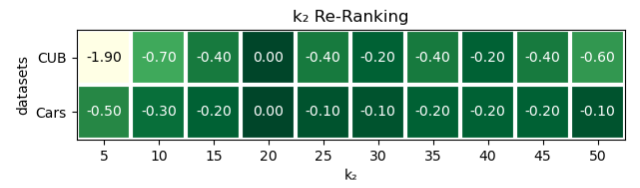}
\vspace{-0.4cm}
\caption{Sensitivity of R@1 with respect to different $k2-$values of re-ranking.}
\vspace{+0.3cm}
\label{fig:k2}
\end{minipage}

\end{figure}

In Fig. \ref{fig:beta}, we show that the performance on \textit{CUB-200-2011} is way more sensitive towards different $\beta-$values than the performance on \textit{Cars 196} mirroring the difference in the datasets. 
While the same holds for the negative slope of the LeakyReLU (see Fig. \ref{fig:leaky}), different combinations of maximum pooling and average pooling are similarly sensitive for both datasets (see Fig. \ref{fig:max}) where the $\alpha-$value represents the weighting between max and average pooling:

\begin{equation}
    f_{flat} = \alpha \; maxpool(f) + (1-\alpha) avgpool(f)
\end{equation}

$f \in \mathcal{R}^{C \times H \times W}$ is a feature map with $C$ channels of height $H$ and width $W$, $maxpool$ and $avgpool$ represent maximum pooling and average pooling, and $f_{flat} \in \mathcal{R}^{C}$ is the flattened feature vector. As can be seen from the sensitivity analysis of the pooling operation, only very large and very low $\alpha-$values lead to a more significant performance drop.

We also investigate the three re-ranking hyperparameters $k1$, $k2$, and $\lambda$ with respect to the performance sensitivity. As can be seen from Fig. \ref{fig:lamda_re_ranking}, the performance gets worse for large and small $\lambda$-values while being highly stable for values between $0.85$ and $0.95$ for both datasets. As can be seen from Fig. \ref{fig:k1}, the performance on \textit{Cars 196} is hardly influenced by the size of the $k_1-$NN set while on \textit{CUB-200-2011}, we see a slight decrease in performance. Similarly, for $k_2$ the performance on \textit{CUB-200-2011} is more influenced than the performance on Cars dataset (see Fig. \ref{fig:k2}).

Overall, the performance is highly robust towards the choice of inference hyperparameters such that the performance only drops drastically for extreme parameter choices.

\section*{Convergence rate.} 
\label{convergence_rate}
In Fig.~\ref{fig:abl_convergence}, we present the convergence rate of the model on the \textit{Cars 196} dataset. Within the first $30$ epochs, our model reaches higher results than the other models, making our model significantly faster than other approaches. The other models except Proxy-NCA \cite{DBLP:conf/iccv/Movshovitz-Attias17}, need hundreds of epochs to converge. 
Additionally, we compare the training time with Proxy-NCA \cite{DBLP:conf/iccv/Movshovitz-Attias17}. On a single Volta V100 GPU, the average running time of our method per epoch is $23.59$ seconds on \textit{CUB-200-2011} and $39.35$ seconds on \textit{Cars 196}, compared to $27.43$ and $42.56$ seconds of Proxy-NCA \cite{DBLP:conf/iccv/Movshovitz-Attias17}.
Hence, our method is faster than one of the fastest methods in the literature. Note, the inference time of every method is the same because the network is only used for feature embedding extraction. 

\section*{Dealing with negative similarities}
\label{negative_similarities}

Equation (6) in the main paper assumes that the matrix of similarity is non-negative.
However, for similarity computation, we use a correlation metric (see Equation (3) in the main paper) which produces values in the range $[-1, 1]$.
In similar situations, different authors propose different methods to deal with the negative outputs.
The most common approach is to shift the matrix of similarity towards the positive regime by subtracting the biggest negative value from every entry in the matrix \cite{DBLP:journals/neco/ErdemP12}.
Nonetheless, this shift has a side effect: If a sample of class $k_1$ has very low similarities to the elements of a large group of samples of class $k_2$, these similarity values, which after being shifted are all positive, will be summed up. If the cardinality of class $k_2$ is very large, then summing up all these small values leads to a large value, and consequently affects the solution of the algorithm.
What we want instead, is to ignore these negative similarities, hence we propose {\it clamping}. More concretely, we use a \textit{ReLU} activation function over the output of Equation (1).

We compare the results of shifting vs clamping using the Group Loss (GL). On the \textit{Cars 196} dataset, we do not see a significant difference between the two approaches.
However, on the \textit{CUB-200-2011} dataset, the Recall@1 metric is $51$ with shifting, much below the $65.5$ obtained when using clamping.
We investigate the matrix of similarities for the two datasets, and we see that the number of entries with negative values for the \textit{CUB-200-2011} dataset is higher than for the \textit{Cars 196} dataset. This explains the difference in behavior, and also verifies our hypothesis that clamping is a better strategy to use within {\it Group Loss}.

\section*{Other backbones}
\label{other_backbones}

In the main paper, we perform all experiments using a GoogleNet backbone with batch normalization. This choice is motivated by the fact that most methods use this backbone, making comparisons fair. 
In this section, we explore the performance of our method for other backbone architectures, to show the generality of our proposed loss formulation. 
We choose to train a few networks from Densenet family \cite{DBLP:conf/cvpr/HuangLMW17}. Densenets are a modern CNN architecture that show similar classification accuracy to BN-Inception in most tasks (so they are a similarly strong classification baseline \footnote{The classification accuracy of different backbones can be found in the following link: \url{https://pytorch.org/docs/stable/torchvision/models.html}. BN-Inception's top 1/top 5 error is 7.8\%/25.2\%, very similar to those of Densenet121 (7.8\%/25.4\%).}). 
Furthermore, by training multiple networks of the same family, we can study the effect of the capacity of the network, i.e., how much can we gain from using a larger network? 
Finally, we are interested in studying if the choice of hyperparameters can be transferred from one backbone to another.

We present the results of our method using Densenet backbones in Tab. \ref{tab:densenets}. 
We use the same hyperparameters as the ones used for the BN-Inception experiments, reaching state-of-the-art results on both \textit{Cars 196} \cite{KrauseStarkDengFei-Fei_3DRR2013} and \textit{Stanford Online Products} \cite{DBLP:conf/cvpr/SongXJS16} datasets, even compared to ensemble and sampling methods. 
The results in \textit{Stanford Online Products} \cite{DBLP:conf/cvpr/SongXJS16} are particularly impressive considering that Group Loss was the first time any method in the literature has broken the $80$ point barrier in Recall@1 metric. 
We also reach state-of-the-art results on the \textit{CUB-200-2011} \cite{WahCUB_200_2011} dataset when we consider only methods that do not use ensembles (with the Group Loss ensemble reaching the highest results in this dataset). 
We observe a clear trend when increasing the number of parameters (weights), with the best results on both \textit{Cars 196} \cite{KrauseStarkDengFei-Fei_3DRR2013} and \textit{Stanford Online Products} \cite{DBLP:conf/cvpr/SongXJS16} datasets being achieved by the largest network, Densenet161 (whom has a lower number of convolutional layers than Densenet169 and Densenet201, but it has a higher number of weights/parameters).

Finally, we study the effects of hyperparameter optimization using the Group Loss (GL). Despite that the networks reached state-of-the-art results even without any hyperparameter tuning, we expect a minimum amount of hyperparameter tuning to help. 
To this end, we used random search \cite{DBLP:journals/jmlr/BergstraB12} to optimize the hyperparameters of our best network on the \textit{Cars 196} \cite{KrauseStarkDengFei-Fei_3DRR2013} dataset. We reach a $90.7$ score ($2pp$ higher score than the network with default hyperparameters) in Recall@1, and $77.6$ score ($3pp$ higher score than the network with default hyperparameters) in NMI metric, showing that individual hyperparameter optimization can boost the performance. 
The score of $90.7$ in Recall@1 is not only by far the highest score ever achieved, but also the first time any method has broken the $90$ point barrier in Recall@1 metric when evaluated on the \textit{Cars 196} \cite{KrauseStarkDengFei-Fei_3DRR2013} dataset.

\begin{table*}[t]
\centering
\resizebox{\textwidth}{!}{
\begin{tabular}{c|c|c|c|c|c|c|c|c|c}
\multicolumn{1}{c}{Model} & \multicolumn{3}{c}{CUB} & \multicolumn{3}{c}{CARS} & \multicolumn{3}{c}{SOP}  \\ 
\hline 
& Params & R@1 & NMI & Params & R@1 & NMI & Params & R@1 & NMI\\ \hline
GL Densenet121                 & 7056356      &   \textbf{65.5}  & 69.4 &  7054306    & 88.1  & 74.2  &  18554806   &   78.2 &   91.5     \\
GL Densenet161                 &  26692900     &  64.7  &  68.7 &  26688482    & \textbf{88.7}  & 74.6 & 51473462    & \textbf{80.3} &  \textbf{92.3}        \\
GL Densenet169                 &  12650980     &  65.4   & \textbf{69.5} &  12647650    & 88.4 & 75.2 & 31328950    &    79.4 &  92.0    \\
GL Densenet201                 &  18285028     &  63.7  & 68.4  &  18281186    &  88.6 & \textbf{75.8}  & 39834806    &    79.8   &  92.1     \\ \hline
GL Inception v2                 &   10845216    &  \textbf{65.5}  & 69.0  &  10846240    & 85.6   & 72.7 & 16589856    & 75.7   & 91.1 \\ \hline
\end{tabular}}
\caption{The results of Group Loss in Densenet backbones.}
\label{tab:densenets}
\end{table*}

\section*{Ensembles}
\label{ensembles}

We do an experiment using the Group Loss (GL), where we build an ensemble of networks trained by our model. Unlike other methods in the literature, we simply train independently $k$ networks and then concatenate their features during inference.
In Table \ref{tab:loss_ensembles} we present the results of our ensemble, and compare them with the results of other ensemble approaches. Our ensemble method (using $5$ neural networks) is the highest performing model in \textit{CUB-200-2011}, outperforming the second-best method (Divide and Conquer \cite{DDBLP:conf/cvpr/Sanakoyeu2019}) by $1pp$ in Recall@1 and by $0.4pp$ in NMI. In \textit{Cars 196} our method outperforms the second best method (ABE 8 \cite{DBLP:conf/eccv/KimGCLK18}) by $2.8pp$ in Recall@1. The second best method in NMI metric is the ensemble version of RLL \cite{DBLP:conf/cvpr/WangHKHGR19} which gets outperformed by $2.4pp$ from the Group Loss. In \textit{Stanford Online Products}, our ensemble reaches the third-highest result on the Recall@1 metric (after RLL \cite{DBLP:conf/cvpr/WangHKHGR19} and GPW \cite{DDBLP:conf/cvpr/Wand2019}) while increasing the gap with the other methods in NMI metric.

\begin{table*}[t]
\centering
    \resizebox{\textwidth}{!}{%
    \begin{tabular}{@{}l|ccccc|ccccc|cccc@{}}

\textbf{} & \multicolumn{5}{c}{CUB-200-2011} & \multicolumn{5}{c}{CARS 196} & \multicolumn{4}{c}{Stanford Online Products}\\ \hline
\textbf{Loss+Ensembles} & R@1 & R@2 & R@4 & R@8 & NMI & R@1 & R@2 & R@4 & R@8 & NMI & R@1 & R@10 & R@100 & NMI  \\ \hline
BIER 6  \cite{DBLP:conf/iccv/OpitzWPB17} & 55.3 & 67.2 & 76.9 & 85.1 &  - & 75.0 & 83.9 & 90.3 & 94.3 &  - &  72.7 & 86.5 & 94.0  & -\\
HDC 3  \cite{DBLP:conf/iccv/YuanYZ17} &  54.6 & 66.8 & 77.6 & 85.9 & - &  78.0 & 85.8 & 91.1 & 95.1 & - & 70.1 & 84.9 & 93.2 & - \\
ABE 2  \cite{DBLP:conf/eccv/KimGCLK18} &  55.7 & 67.9 & 78.3 & 85.5 & - & 76.8 & 84.9 & 90.2 & 94.0 & - & 75.4 & 88.0 & 94.7  & -  \\
ABE 8  \cite{DBLP:conf/eccv/KimGCLK18} &  60.6 & 71.5 & 79.8 & 87.4 & - & 85.2 & 90.5 & 94.0 & 96.1 &  - & 76.3 & 88.4 &  94.8 & - \\
A-BIER 6 \cite{DBLP:journals/corr/abs-1801-04815} & 57.5 & 68.7 & 78.3 & 86.2 & - &  82.0 & 89.0 & 93.2 & 96.1 &  - & 74.2 & 86.9 & 94.0 & - \\
D and C 8  \cite{DDBLP:conf/cvpr/Sanakoyeu2019} & 65.9 & 76.6 & 84.4 & 90.6 & 69.6 & 84.6 & 90.7 & 94.1 & 96.5 &  70.3 &   75.9   & 88.4 & 94.9 &90.2 \\ 
RLL 3  \cite{DBLP:conf/cvpr/WangHKHGR19} & 61.3 & 72.7 & 82.7 & 89.4 & 66.1 & 82.1 & 89.3 & 93.7 & 96.7 & 71.8 &  \textbf{79.8} & \textbf{91.3} & \textbf{96.3} & 90.4\\
\hline
\textbf{Group Loss 2-ensemble} & 65.8 & 76.7 & 85.2 & 91.2 & 68.5 &  86.2 & 91.6 & 95.0 & 97.1 & 72.6 & 75.9 & 88.0 & 94.5 & \textbf{91.1}\\
\textbf{Group Loss++ 5-ensemble} &  \textbf{66.9} & \textbf{77.1} & \textbf{85.4} & \textbf{91.5} & \textbf{70.0} &  \textbf{88.0} & \textbf{92.5} & \textbf{95.7} & \textbf{97.5} & \textbf{74.2} &  76.3 & 88.3 & 94.6 & \textbf{91.1} \\ 
\end{tabular}}
\caption{Retrieval and Clustering performance of our ensemble compared with ensemble methods. Bold indicates best results. }
\label{tab:loss_ensembles}
\end{table*}

\section*{Temperature scaling}
\label{temperature}
We mentioned in the main paper that as input to the Group Loss (step 3 of the algorithm) we initialize the matrix of priors $X(0)$ from the softmax layer of the neural network. 
Following the works of \cite{DBLP:conf/icml/GuoPSW17,DBLP:conf/nips/BerthelotCGPOR19,DBLP:journals/corr/abs-1811-12649}, we apply a sharpening function to reduce the entropy of the softmax distribution. We use the common approach of adjusting the {\it temperature} of this categorical distribution, known as temperature scaling. 
Intuitively, this procedure calibrates our network and in turn, provides a  more informative prior to the dynamical system. Additionally, this calibration allows the dynamical system to be more effective in adjusting the predictions, i.e, it is easier to change the probability of a class if its initial value is $0.6$ rather than $0.95$. The function is implemented using the following equation:

\begin{equation}
    T_{softmax}(z_i) = \frac{e^{z_i/T}}{\sum_i{e^{z_i/T}}},
\end{equation}

which can be efficiently implemented by simply dividing the prediction logits by a constant $T$.

Recent works in supervised learning \cite{DBLP:conf/icml/GuoPSW17} and semi-supervised learning \cite{DBLP:conf/nips/BerthelotCGPOR19} have found that temperature calibration improves the accuracy for the image classification task. We arrive at similar conclusions for the task of metric learning, obtaining $2.5pp$ better Recall@1 scores on \textit{CUB-200-2011} \cite{WahCUB_200_2011} and $2pp$ better scores on \textit{Cars 196} \cite{KrauseStarkDengFei-Fei_3DRR2013}.
Note, the methods of Table 1 (main paper) that use a classification loss, use also temperature scaling.

\section*{Qualitative results}
\label{qualitative}

\begin{figure*}[ht]
\centering
\begin{minipage}{.333\textwidth}
  \flushleft 
\includegraphics[width=\textwidth, trim={0.5cm 4cm 12cm 0.1cm},clip]{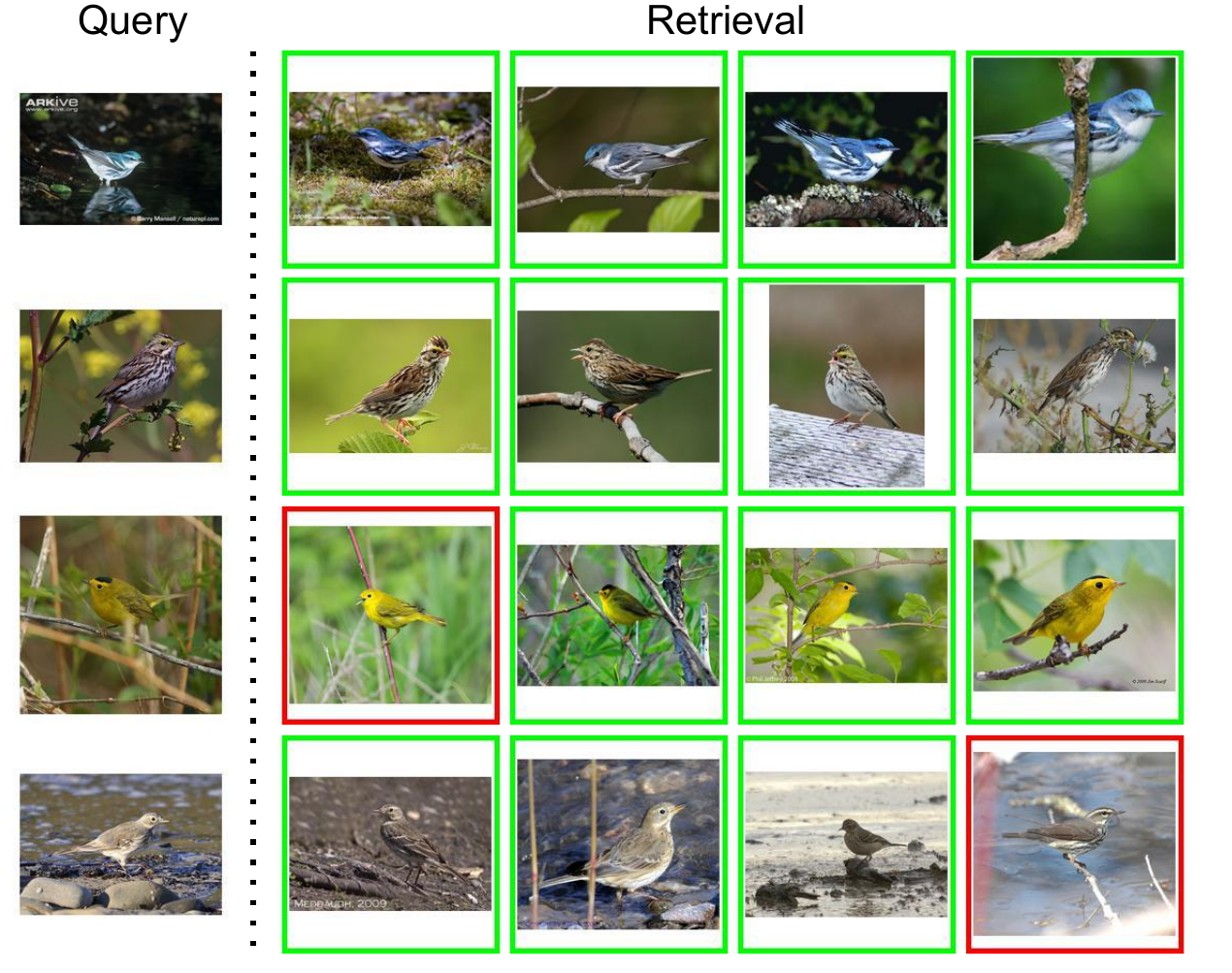}
\end{minipage}%
\begin{minipage}{.333\textwidth}
  \centering
\includegraphics[width=\textwidth, trim={0.5cm 4cm 12cm 0.1cm},clip]{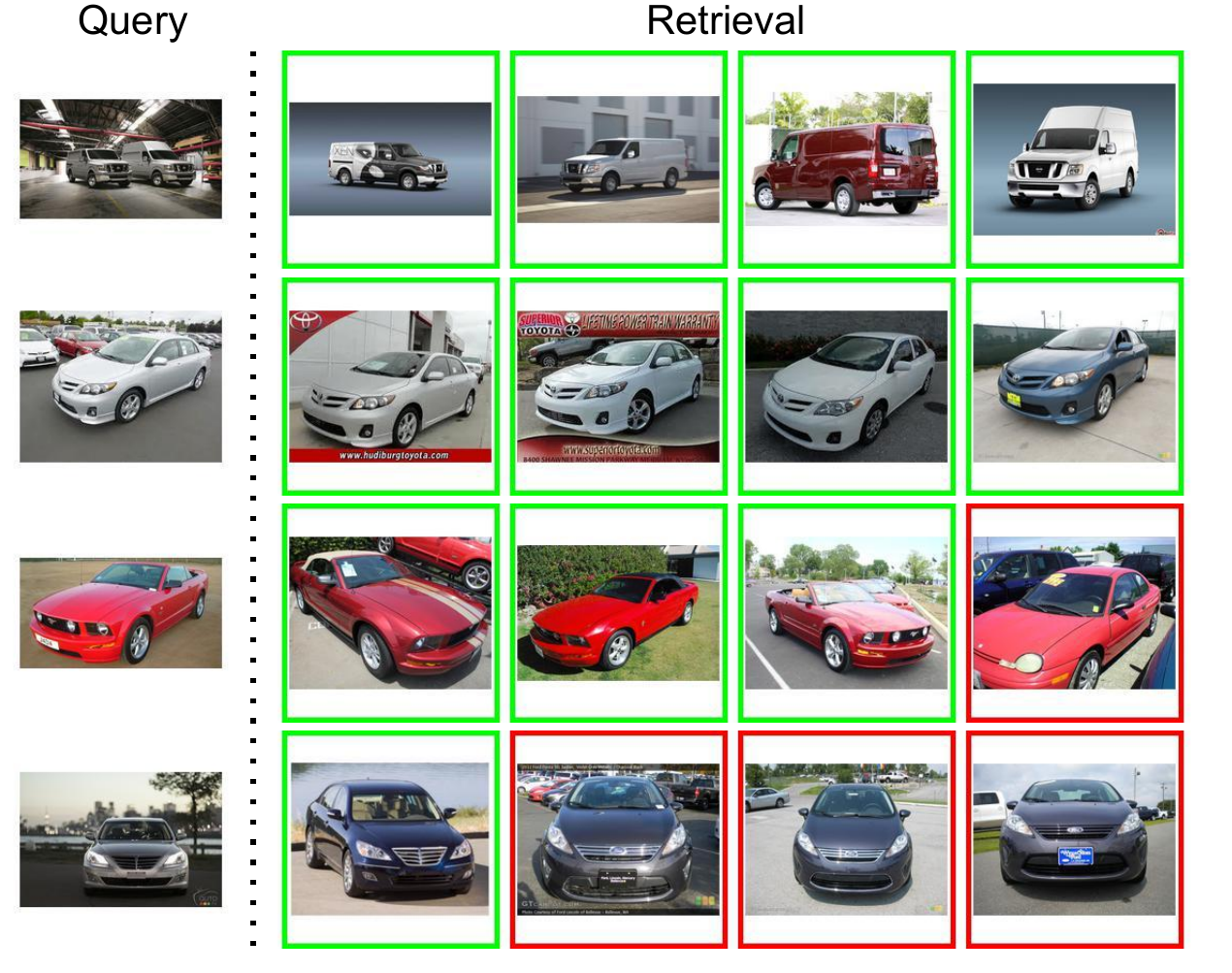}
\end{minipage}%
\begin{minipage}{.333\textwidth}
  \flushright 
\includegraphics[width=\textwidth, trim={0.5cm 4cm 12cm 0.1cm},clip]{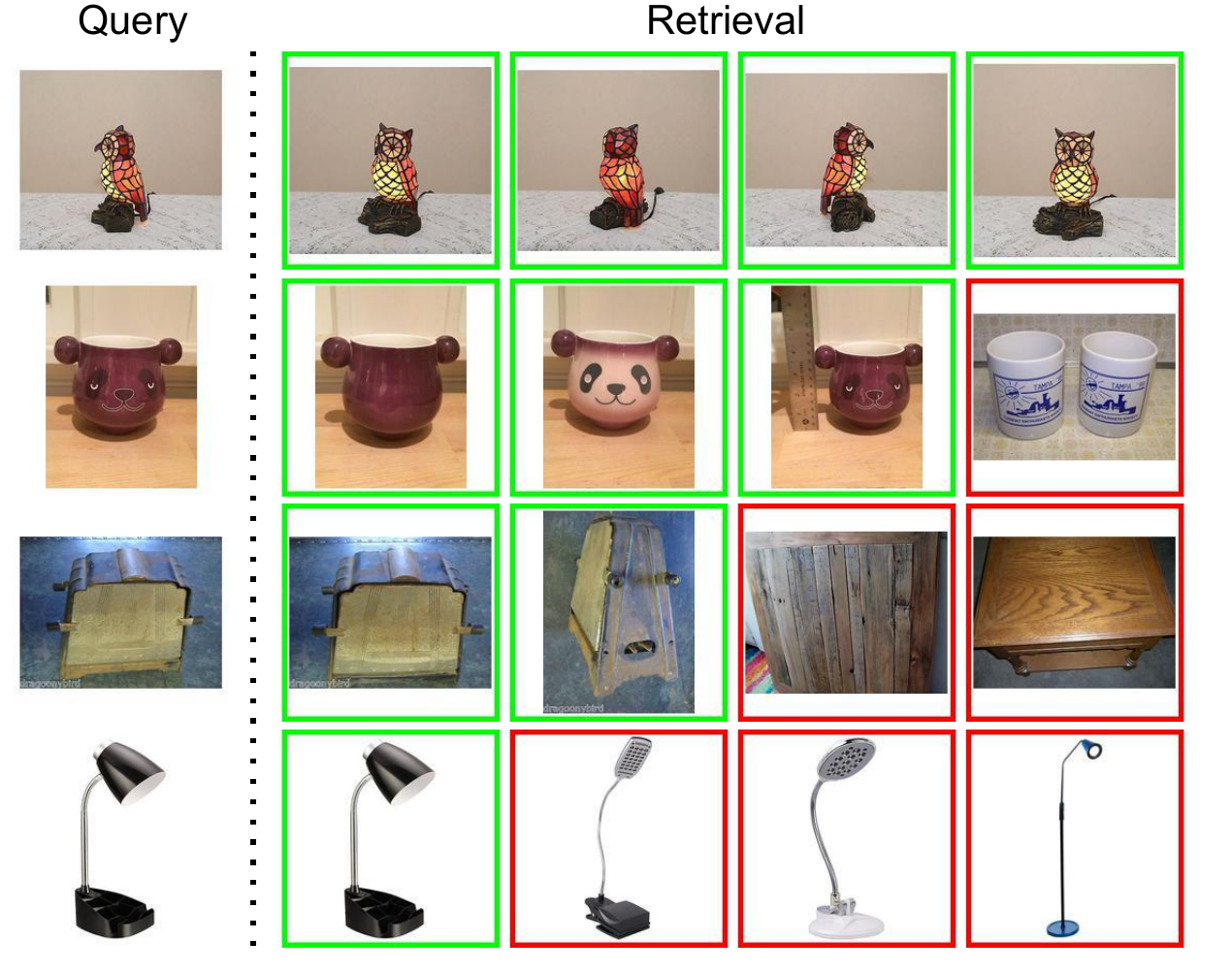}
\end{minipage}

\caption{Retrieval results on a set of images from the
\textit{CUB-200-2011} (left), \textit{Cars 196} (middle), and \textit{Stanford Online Products} (right) datasets using our Group Loss model. The left column contains query images. The results are ranked by distance. The green square indicates that the retrieved image is from the same class as the query image, while the red box indicates that the retrieved image is from a different class.}  
\label{fig:retrieval}
\end{figure*}

In Fig. \ref{fig:retrieval} we present qualitative results on the retrieval task in \textit{CUB-200-2011}, \textit{Cars 196} and \textit{Stanford Online Products} datasets. In all cases, the query image is given on the left, with the four nearest neighbors given on the right. 
Green boxes indicate the cases where the retrieved image is of the same class as the query image, and red boxes indicate a different class. 
As we can see, our model is able to perform well even in cases where the images suffer from occlusion and rotation. On the \textit{Cars 196} dataset, we see a successful retrieval even when the query image is taken indoors and the retrieved image outdoors, and vice-versa. 
The first example of \textit{Cars 196} dataset is of particular interest. Despite that the query image contains $2$ cars, its four nearest neighbors have the same class as the query image, showing the robustness of the algorithm to uncommon input image configurations. 

\begin{figure*}[ht]
\centering
\includegraphics[width=0.95\textwidth, trim={1cm 3cm 0.5cm 5cm},clip]{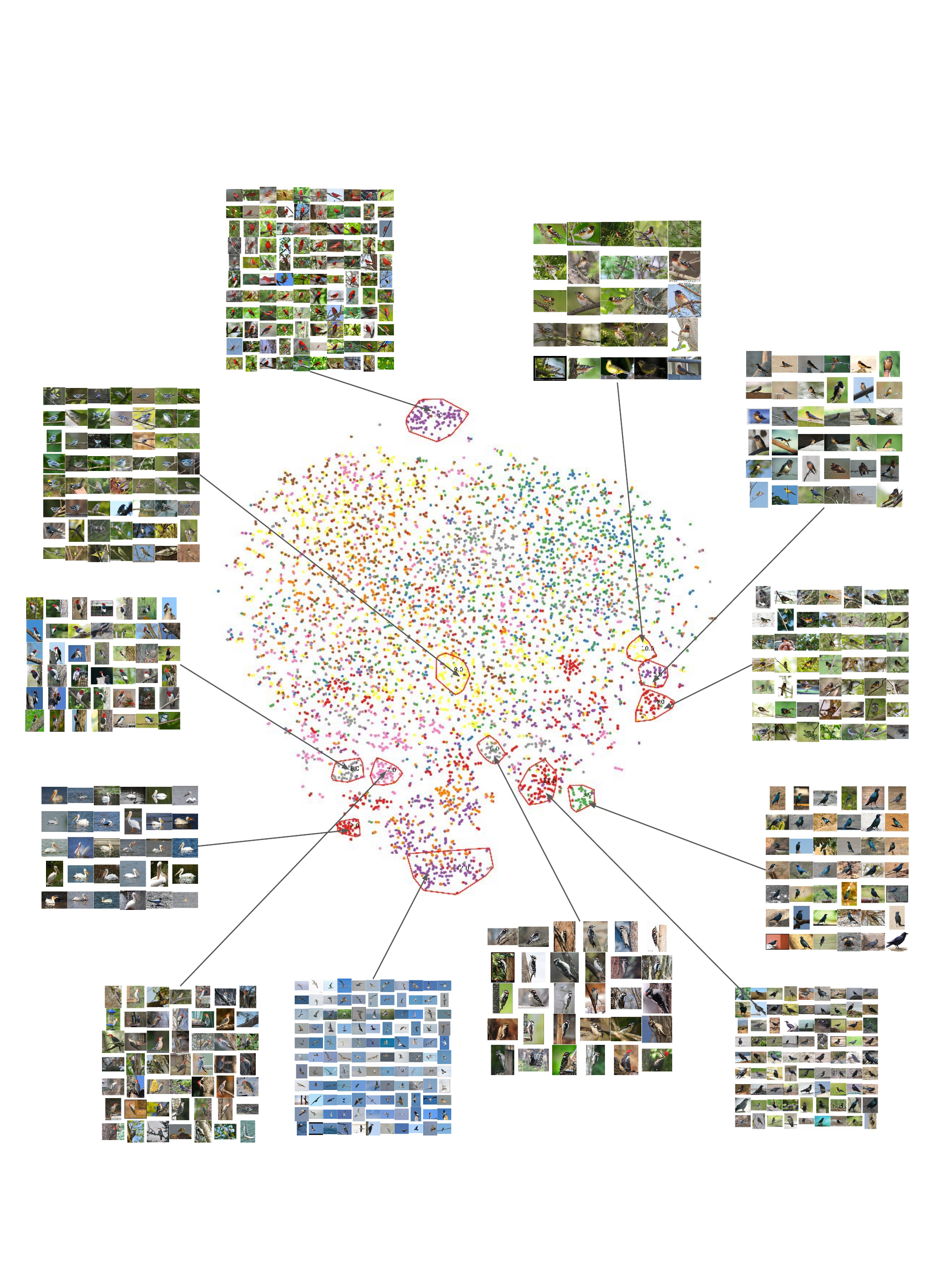}
  \caption{t-SNE \cite{DBLP:journals/ml/MaatenH12} visualization of our embedding
on the CUB-200-2011 \cite{WahCUB_200_2011} dataset, with some clusters highlighted. Best viewed on a monitor
when zoomed in.}
  \label{fig:t-sne}
\end{figure*}

In Fig. \ref{fig:t-sne}, we visualize the t-distributed stochastic neighbor embedding (t-SNE) \cite{DBLP:journals/ml/MaatenH12} of the embedding vectors obtained by our method on the \textit{CUB-200-2011} \cite{WahCUB_200_2011} dataset. 
The plot is best viewed on a high-resolution monitor when zoomed in. We highlight several representative groups by enlarging the corresponding regions in the corners. Despite the large pose and appearance variation, our method efficiently generates a compact feature mapping that preserves semantic similarity.

\ifCLASSOPTIONcaptionsoff
  \newpage
\fi



%
\bibliographystyle{IEEEtran}
\bibliography{bib_short.bib}

%

\vspace{-1.1cm}
\begin{IEEEbiography}[{\includegraphics[width=1in,height=1.25in,clip,keepaspectratio]{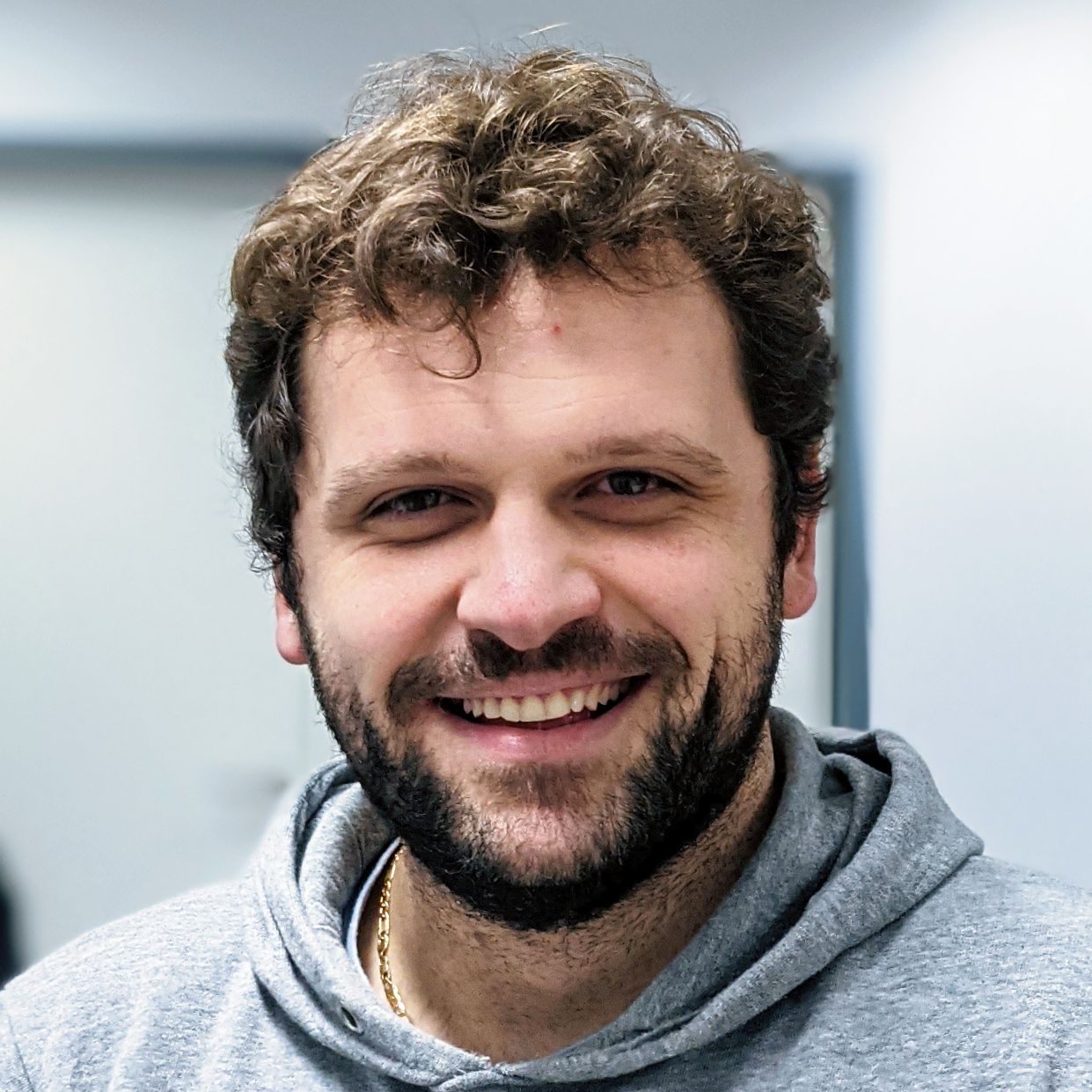}}]{Dr. Ismail Elezi}
obtained his Ph.D. at Ca' Foscari University of Venice under the supervision of Prof. Marcello Pelillo and Prof. Thilo Stadelmann. In 2019, he was a visiting Ph.D. student at the Dynamic Vision and Learning (DVL) Group headed by Prof. Laura Leal-Taixe working on metric learning and generative adversarial networks. He then did an internship at Nvidia (Santa Clara) working on active learning. Since October 2020, he is a postdoctoral researcher at DVL, working on topics such as metric learning, generative adversarial networks, active learning, and semi-supervised learning. He recently received a Humboldt Research Fellowship.
\end{IEEEbiography}

\vspace{-0.3cm}
\begin{IEEEbiography}[{\includegraphics[width=1in,height=1.25in,clip,keepaspectratio]{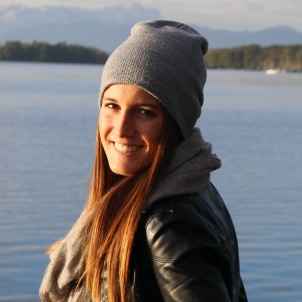}}]{Jenny Seidenschwarz}
obtained her bachelor's degree in Mechanical Engineering at the Technical University of Munich, and her master's degree in Robotics, Cognition, and Intelligence also at the Technical University of Munich. In her Master's, she specialized in Deep Learning and Machine Learning and wrote her thesis on Deep Metric Learning at the Dynamic Vision and Learning group under the supervision of Dr. Ismail Elezi and Prof. Dr. Laura Leal-Taixé. Following her thesis, she started a Ph.D. position at the beginning of 2021 also at the Dynamic Vision and Learning group.
\end{IEEEbiography}

\begin{IEEEbiography}[{\includegraphics[width=1in,height=1.25in,clip,keepaspectratio]{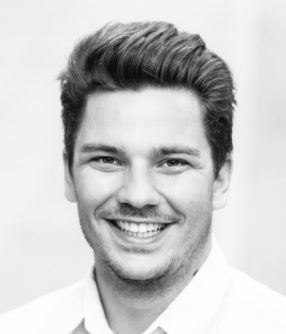}}]{Laurin Wagner}
obtained his bachelor's degree in Mathematics at the Technical University of Munich, and his Master's degree in Data Science at the Technical University of Munich and the Royal Institute of Technology in Stockholm. In his Master's, he focused on Machine Learning and specialized in Computer Vision. He wrote his  thesis in the area of Deep Metric Learning at the Dynamic Vision and Learning group under the supervision of Dr. Ismail Elezi and Prof. Dr. Laura Leal-Taixé. 
\end{IEEEbiography}

\vspace{-1.25cm}
\begin{IEEEbiography}[{\includegraphics[width=1in,height=1.25in,clip,keepaspectratio]{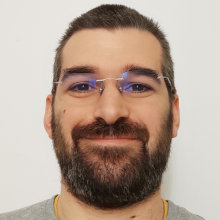}}]
{Dr. Sebastiano Vascon} is an assistant professor at Ca' Foscari University of Venice. In 2016 he obtained the title of Ph.D. from the Italian Institute of Technology under the supervision of prof. Vittorio Murino. He then moved to University of Venice working as a PostDoc within the group of prof. Marcello Pelillo. His research activities are in the fields of computer vision, game theory, machine, and deep learning.
\end{IEEEbiography}

\vspace{-1.25cm}
\begin{IEEEbiography}[{\includegraphics[width=1in,height=1.25in,clip,keepaspectratio]{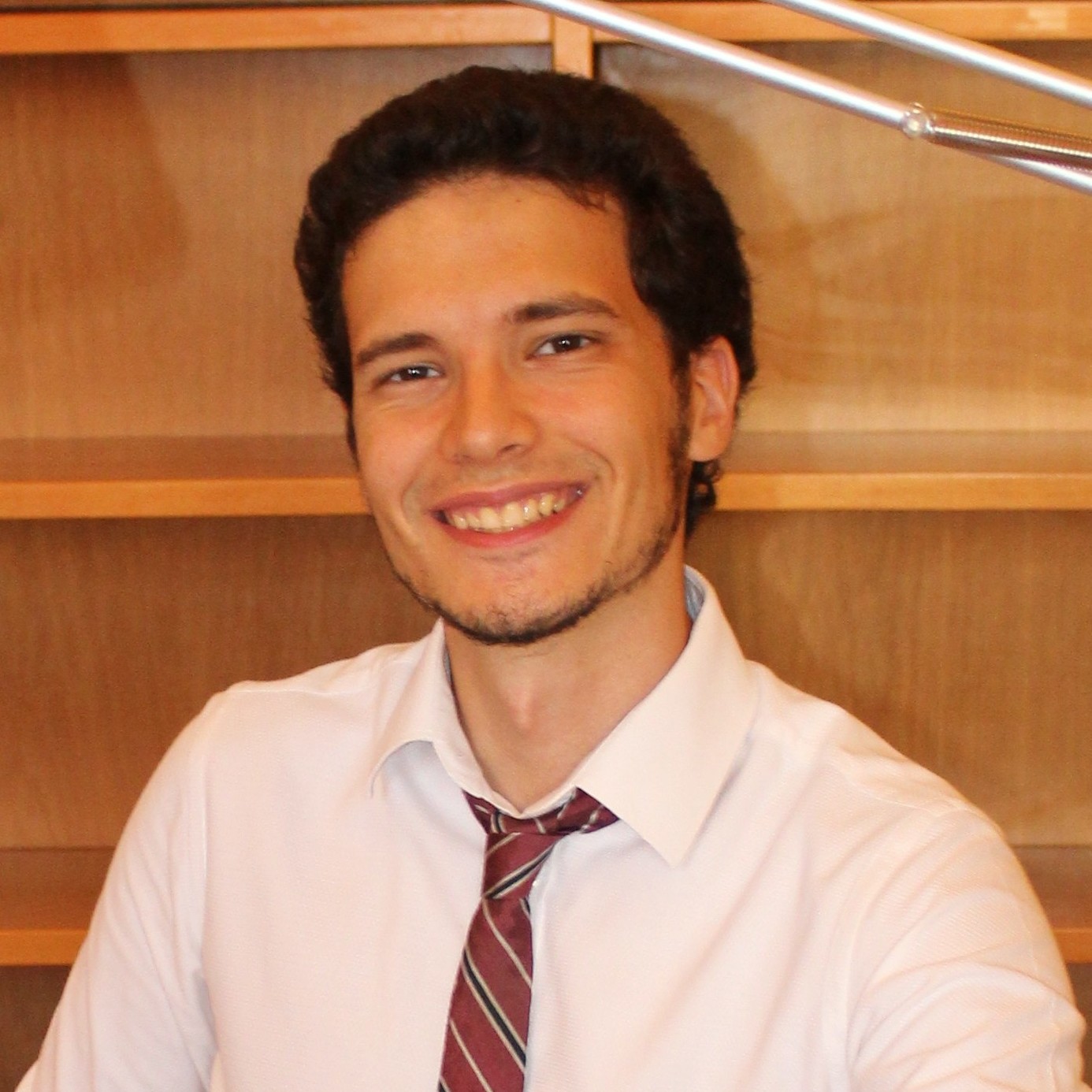}}]
{Dr. Alessandro Torcinovich} is a postdoc at Ca' Foscari University of Venice. In 2021, he obtained his doctoral degree under the supervision of prof. Marcello Pelillo. During his Ph.D. program, he worked on the integration of game-theoretic techniques and deep learning methods for addressing weakly supervised tasks. In addition, he brought on a technology transfer project in collaboration with the ESA on anomaly detection tasks. 
\end{IEEEbiography}

\vspace{-1.25cm}
\begin{IEEEbiography}[{\includegraphics[width=1in,height=1.25in,clip,keepaspectratio]{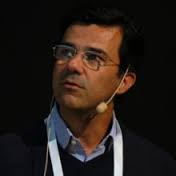}}]
{Prof. Marcello Pelillo} is a full professor of Computer Science at Ca’ Foscari University in Venice, Italy,
where he directs the Computer Vision
and Pattern Recognition group. He held visiting
research positions at Yale University, McGill University, the University of Vienna, York University
(UK), the University College London, the National ICT Australia (NICTA). He has been General Chair for ICCV 2017, Program Chair for ICPR 2020, and has founded many conferences and workshops (e.g., EMMCVPR, SIMBAD, IWCV). He serves (has served) on the Editorial Boards
of the journals IEEE Transactions on Pattern Analysis and Machine
Intelligence (PAMI), Pattern Recognition, IET Computer Vision, Frontiers
in Computer Image Analysis, Brain Informatics, and serves on the
Advisory Board of the International Journal of Machine Learning and
Cybernetics. He has been elected a Fellow of the IEEE and a Fellow of the IAPR, and has recently been appointed IEEE SMC Distinguished Lecturer. His Erdos number is 2. 
\end{IEEEbiography}

\vspace{-1.25cm}
\begin{IEEEbiography}[{\includegraphics[width=1in,height=1.25in,clip,keepaspectratio]{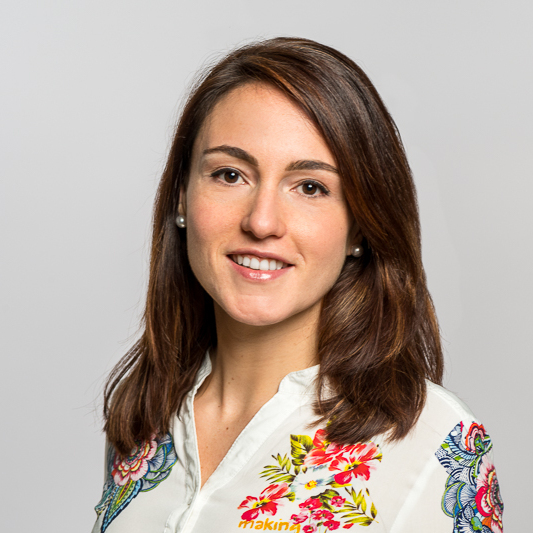}}]{Prof. Dr. Laura Leal-Taixé}
is a tenure-track professor (W2) at the Technical University of Munich leading the Dynamic Vision and Learning group and a Principal research scientist at Argo AI. Before that, she spent two years as a postdoctoral researcher at ETH Zurich, Switzerland, and a year as a senior postdoctoral researcher in the Computer Vision Group at the Technical University in Munich. She obtained her Ph.D. from the Leibniz University of Hannover in Germany, spending a year as a visiting scholar at the University of Michigan, Ann Arbor, USA. She pursued B.Sc. and M.Sc. in Telecommunications Engineering at the Technical University of Catalonia (UPC) in her native city of Barcelona. She is a recipient of the Sofja Kovalevskaja Award of 1.65 million euros for her project socialMaps.
\end{IEEEbiography}





\end{document}